\documentclass[runningheads]{llncs}
\usepackage{graphicx}
\usepackage{amsmath,amssymb} 
\usepackage{color}
\usepackage{times}
\usepackage{epsfig}
\usepackage{amsmath}
\usepackage{amssymb}
\usepackage{mathptmx}
\usepackage{array}
\usepackage{multirow}
\usepackage{booktabs}
\usepackage{subcaption}
\usepackage{caption}

\usepackage[breaklinks=true]{hyperref}
\usepackage{breakcites}

\graphicspath{{figures/}}
\newlength\figwidth

\usepackage[normalem]{ulem}
\useunder{\uline}{\ul}{}

\long\def\ignorethis#1{}

\newcommand\para[1]{\noindent{#1}}

\def\be {\begin{equation}}
\def\ee {\end{equation}}
\def\beas {\begin{eqnarray*}}
	\def\eeas {\end{eqnarray*}}
\def\bea {\begin{eqnarray}}
\def\eea {\end{eqnarray}}
\def\bes {\begin{equation*}}
\def\ees {\end{equation*}}
\def\ba {\begin{align}}
\def\ea {\end{align}}
\def\barr {\begin{array}}
	\def\earr {\end{array}}

\newcommand{\bx}{{\mathbf{x}}}

\newcommand{\bm}{{\mathbf{m}}}
\newcommand{\bb}{{\mathbf{b}}}

\makeatletter
\usepackage{xspace}
\def\@onedot{\ifx\@let@token.\else.\null\fi\xspace}
\DeclareRobustCommand\onedot{\futurelet\@let@token\@onedot}

\newcommand{\figref}[1]{Figure~\ref{#1}} 

\newcommand{\secref}[1]{Section~\ref{#1}}
\newcommand{\tabref}[1]{Table~\ref{#1}}

\def\eg{\emph{e.g}\onedot} 
\def\ie{\emph{i.e}\onedot} 
 
 \def\vs{\emph{vs}\onedot}
\def\wrt{w.r.t\onedot} 
\def\etal{\emph{et al}\onedot}

\makeatother

\newcommand{\tb}[1]{\textbf{#1}}

\newcommand{\nd}{\textsuperscript}

\begin{document}
\title{VideoMatch: Matching based Video Object Segmentation} 

\titlerunning{VideoMatch: Matching based Video Object Segmentation}
%
\author{Yuan-Ting Hu$^{1}$ \and
Jia-Bin Huang$^{2}$ \and
Alexander G. Schwing$^{1}$}
%
\authorrunning{Y.-T. Hu, J.-B. Huang  and A. G. Schwing}
%

\institute{
$^1$University of Illinois at Urbana-Champaign 
\hspace{20pt}
$^2$Virginia Tech \\
\hspace{10pt}
\email{\{ythu2,aschwing\}@illinois.edu}  \hspace{10pt}
\email{jbhuang@vt.edu}
}
\maketitle              
%
\begin{abstract}
Video object segmentation is challenging yet important in a wide variety of applications for video analysis. Recent works formulate video object segmentation as a prediction task using deep nets to achieve appealing state-of-the-art performance. Due to the formulation as a prediction task, most of these methods require  fine-tuning during test time, such that the deep nets memorize the appearance of the objects of interest in the given video. However, fine-tuning is time-consuming and computationally expensive, hence the algorithms are far from real time. To address this issue, we develop a novel matching based algorithm for video object segmentation. In contrast to memorization based classification techniques, the proposed approach learns to match extracted features to a provided template without memorizing the appearance of the objects. We validate the effectiveness and the robustness of the proposed method on the challenging DAVIS-16, DAVIS-17, Youtube-Objects and JumpCut datasets. Extensive results show that our method achieves comparable performance without fine-tuning and  is  much more favorable in terms of computational time. 
\end{abstract}

\section{Introduction}
\label{sec:intro}

Video segmentation plays a pivotal role in a wide variety of applications ranging from object identification, video editing to video compression. Despite the fact that delineation and tracking of objects are seemingly trivial for humans in many cases, video object segmentation remains challenging for algorithms due to occlusions, fast motion, motion blur, and significant appearance variation over time. 

Research efforts developing effective techniques for video object segmentation continue to grow, partly because of the recent release of high-quality datasets, \eg, the DAVIS dataset~\cite{PerazziCVPR16,pont2017DAVIS}. 
Two of the main setups for video object segmentation are the unsupervised and the semi-supervised setting~\cite{PerazziCVPR16,pont2017DAVIS}. Both cases are analogous in that the semantic class of the objects to be segmented during testing are not known ahead of time. Both cases differ in the supervisory signal that is available at test time. While no supervisory signal is available during testing in the unsupervised setting, the ground truth segmentation mask of the first frame is assumed to be known in the semi-supervised case. 
With video editing applications in mind, here, we focus on the {\em semi-supervised setting}, \ie, our goal is to delineate in all frames of  the video the object of interest which is specified in the first frame of the video. 

Taking advantage of the provided groundtruth for the first frame, existing semi-supervised video object segmentation techniques follow deep learning based methods
%
~\cite{caelles2016one,PerazziCVPR17,caelles2017semantically,khoreva2017lucid,tokmakov2017learning1,tokmakov2017learning2,voigtlaender2017online,yoon2017pixel} and fine-tune a pre-trained classifier 
on the given ground truth in the first frame during online testing~\cite{caelles2016one,PerazziCVPR17,JampaniCVPR17,caelles2017semantically,khoreva2017lucid,voigtlaender2017online}. This online fine-tuning of a classifier during testing has been shown to improve accuracy significantly. However, fine-tuning during testing is necessary for each object of interest given in the first frame, takes a significant amount of time, and requires specialized hardware in the form of a very recent GPU due to the memory needs of back-propagation for fine-tuning. 

In contrast, in this paper, we propose a novel end-to-end trainable approach for fast semi-supervised video object segmentation that does not require any fine-tuning. Our approach is based on the intuition that features of the foreground and background in any frame should match features of the foreground and background in the first frame. To ensure that the proposed approach can cope with appearance and geometry changes, we use a deep net to learn the features that should match and adapt the sets of features as inference progresses. 




Our method yields competitive results while saving computational time and memory when compared to the current state-of-the-art approaches. On the recently released DAVIS-16 dataset~\cite{PerazziCVPR16}, our algorithm achieves $81.03\%$ in IoU (intersection over union) while  reducing the running time by one order of magnitude compared to the state-of-the-art, requiring on average only $0.32$ seconds per frame. 


\vspace*{-0.1cm}
\section{Related Work}
\label{sec:related}

Video object segmentation has been extensively studied in the past~\cite{TsaiBMVC2010,LiICCV13,NagarajaICCV15,PriceICCV09,LezamaCVPR11,LeeICCV11,PapazoglouICCV13,XiaoCVPR16,GrundmannCVPR2010,TsaiCVPR2016,jain2017fusionseg,caelles2016one,PerazziCVPR17}. In the following, we first discuss the related literature, (1) focusing on semi-supervised video object segmentation, and (2) discussing unsupervised video object segmentation. Subsequently, we examine the relationship of our work and the tracking and matching literature.

\para{\bf Semi-supervised video object segmentation:} 
Semi-supervised video object segmentation assumes that the groundtruth of the first frame is available during testing. Many approaches in this category employ fine-tuning during testing in order to achieve better performance~\cite{caelles2016one,PerazziCVPR17,JampaniCVPR17,caelles2017semantically,khoreva2017lucid,voigtlaender2017online,hu2017maskrnn,ChenICCV17,LiCVPRW2017}. It has been shown that fine-tuning on the first frame significantly improves accuracy. However, the fine-tuning step is computationally demanding, 
adding more than 700 seconds per video to test time~\cite{caelles2016one}. 

Additional cues such as optical flow~\cite{PerazziCVPR17,LiCVPRW2017,khoreva2017lucid,ChenICCV17}, semantic segmentation~\cite{caelles2017semantically,khoreva2017lucid} and re-identification modules~\cite{LiCVPRW2017} can be integrated into the framework to further improve the accuracy. Since fine-tuning is still required, those cues   increase the computational needs. 

Among the semi-supervised video object segmentation methods, the approach by Yoon~\etal~\cite{yoon2017pixel} is most related to our approach. Yoon~\etal~\cite{yoon2017pixel} also address video object segmentation by pixel matching. Their approach concatenates the features extracted from the template and the input images, and uses fully connected layers to simulate matching between the two images. Importantly, the approach still requires fine-tuning. In addition, the fully connected layers restrict the method to process frames at a specific, pre-defined spatial resolution.

Concurrent to our work, several recent methods (all developed independently) have been proposed to improve the speed of video object segmentation through part-based tracking~\cite{cheng2018fast}, pixel-wise metric learning~\cite{chen2018blazingly}, or network modulation~\cite{yang2018efficient,oh2018fast}. We refer the readers to these works for a more complete picture.

\para{\bf Unsupervised video object segmentation:}
Neither groundtruth nor user annotation is available in the unsupervised video object segmentation setting. Therefore, the unsupervised setup requires algorithms to automatically discover the salient objects in video. Different methods such as motion analysis~\cite{PapazoglouICCV13}, trajectory clustering~\cite{OchsPAMI2014}, and saliency-based spatio-temporal propagation~\cite{FaktorBMVC14,hu2018unsupseg} have been proposed to identify the foreground objects. More recently, deep net based approaches have been discussed~\cite{tokmakov2017learning1,tokmakov2017learning2,jain2017fusionseg}. 

\para{\bf  Object tracking:} 
Semi-supervised video object segmentation and object tracking~\cite{yilmaz2006object,VOT_TPAMI} are related to our approach as they both keep track of the objects through the entire video. However, the two tasks differ in the format of the output. The output of video object segmentation is a pixel-level segmentation mask while the output of object tracking is a bounding box that delineates the position and scale of the object. From the tracking literature, work by Bertinetto~\etal~\cite{bertinetto2016fully} is in a spirit similar to our proposed approach as they formulate tracking by matching. However, due to the difference in the output, Bertinetto~\etal~\cite{bertinetto2016fully} calculated correlation by convolving the whole patch with the given template, while we propose a soft matching for pixel-wise segmentation.

\para{\bf Matching:}
Image matching~\cite{Lowe04,Hu15} has been extensively studied over the last few decades. With the success of deep learning, research focus moved from matching using handcrafted features~\cite{Mikolajczyk05b} to deep features~\cite{yang2017deepcd}. Correlation between the extracted feature maps is typically computed to find correspondences~\cite{Revaud2016}, to estimate optical flow fields~\cite{dosovitskiy2015flownet} and geometric transformations~\cite{rocco2017convolutional}. Since the objective of matching is to find point-to-point correspondences, the result will be noisy if the matching algorithm is directly applied to segmentation. To deal with the noisy prediction, we proposed a soft matching mechanism which estimates the similarity score between different segments as discussed next.
\vspace{-0.5cm}
\section{Matching based Video Object Segmentation}
\vspace{-0.1cm}

In the following, we describe details of the proposed algorithm for video object segmentation. We first formally define the problem setting and provide an overview of our approach in~\secref{sec:Overview}. We then detail the new proposed soft matching mechanism in~\secref{sec:SoftMatching}. Subsequently, we show in~\secref{sec:OnlineUpdate} how our model accommodates appearance changes of objects over time during online testing without the need for finetuning. Finally, we demonstrate how to easily extend our method to instance-level video object segmentation in~\secref{sec:Instance}.

\begin{figure*}[t]
\centering

\includegraphics[width=4.6in]{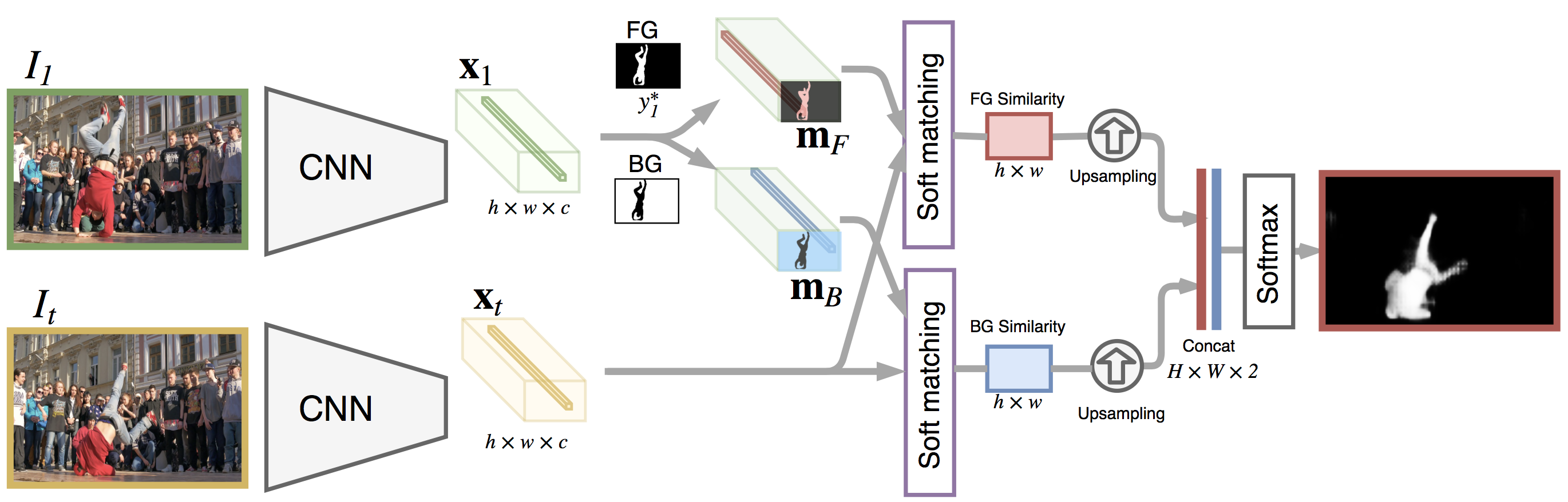}\\

\caption{\tb{Overview of the proposed video object segmentation algorithm.} We use the provided ground truth mask of the first frame to obtain the set of foreground and background features ($\bm_F$ and $\bm_B$). After extracting the feature tensor $\mathbf{x}_t$ from the current frame, we use the proposed soft matching layer to produce FG and BG similarity. We then cancatenate the two similarity scores and generate the final prediction via softmax.
}
\label{fig:Overview}

\end{figure*}
\subsection{Overview}
\label{sec:Overview}

Given a sequence of $T$ video frames $\{I_1, \ldots, I_T\}$ and the groundtruth segmentation $y_1^\ast\in\{1,\ldots,N\}^{W\times H}$ of the first frame $I_1$, the task of semi-supervised video object segmentation is to predict the segmentation masks of the subsequent video frames $I_2, \ldots, I_T$, denoted as $y_2, \ldots,y_T\in \{1,\ldots,N\}^{W\times H}$. Hereby, $N$ is the number of objects of interest in the given video. We denote width and height of the frames as $W$ and $H$. We start by discussing the single instance case ($N = 1$) and explain how to extend the proposed method to $N>1$ in~\secref{sec:Instance}. Importantly, we emphasize that semi-supervised video object segmentation requires object independent formulations since we do not know ahead of time the semantic class of the object to be segmented. 

As the object category and appearance are unknown before test time, a network detecting objectness is usually trained offline. During test time a natural way is to use the given groundtruth for the first frame, \ie $y_1^\ast$, as training data to fine-tune the pretrained objectness network~\cite{caelles2016one,PerazziCVPR17,JampaniCVPR17,caelles2017semantically,khoreva2017lucid,voigtlaender2017online}. Fine-tuning encourages the network to memorize appearance of the object of interest. In previous works on instance-level segmentation, memorization is achieved by fine-tuning a pretrained network $N$ times, \ie, to obtain one fine-tuned network for each object. As discussed before, although this fine-tuning step is the key to improving performance, it introduces a significant amount of processing time overhead and consumes more memory during testing even when there is only one object of interest in the video.

Our idea for efficient video object segmentation is to develop a network which is general enough such that the fine-tuning step can be omitted. To this end, we propose to match features obtained from the test frame $I_t$ to features of the groundtruth foreground and background in the first frame $I_1$ (template). We designed an end-to-end trainable deep neural net, not only to extract features from video frames, but also to match two sets of features. 

To achieve this goal, as shown in \figref{fig:Overview}, we use a Siamese architecture 
that employs a convolutional neural network to compute the two feature maps. We use $\bx_1\in\mathbb{R}^{h\times w\times c}$ and $\bx_t\in\mathbb{R}^{h\times w\times c}$ to refer to feature tensors extracted from the first frame (template) $I_1$ and the test frame $I_t$, respectively. The feature tensors $\bx_1$ and $\bx_t$ are of size $h\times w\times c$, where $c$ is the number of the feature channels and $w$, $h$ are the width and height of the feature maps, proportional to the $W\times H$ sized  video frame. The ratio between $W$ and $w$ depends on the downsampling rate of the convolutional neural net.

Next we define a set of features for the foreground and the background. We refer to those sets via $\bm_F$ and $\bm_B$ respectively. To formally define those sets of features, let $\bx_t^i$ denote the $c$-dimensional vector representing the feature at pixel location $i$ in the downsampled image. Given the groundtruth template $y^\ast_1$ for the first frame, we collect the foreground features $\bm_F$ and background features $\bm_B$ for this first frame via
$$
\bm_F = \{\bx_1^i : i \in g(y_1^\ast)\}\quad\quad\text{ and }\quad\quad \bm_B = \{\bx_1^i : i \notin g(y_1^\ast)\}.
$$
Hereby $g(y_1^\ast)$ is the set of pixels that belongs to foreground as indicated by the ground truth mask $y_1^\ast$ downsampled to size $w\times h$.

After having extracted the foreground ($\bm_F$) and background ($\bm_B$) features from the template and after having computed features $\bx_t\in\mathbb{R}^{h\times w\times c}$ from frame $I_t$ using the same deep net, we match $\bx_t^i$ $\forall i\in\{1, \ldots, wh\}$ to features collected in both sets $\bm_F$ and $\bm_B$ via a soft matching layer. The result of the soft matching layer for each pixel $i$ is its foreground and background matching scores. Subsequently, the foreground and background matching scores are upsampled and normalized into a predicted foreground probability $y_t$ via the softmax operation. We visualize this process in \figref{fig:Overview} and describe the proposed soft matching layer subsequently in greater detail. 

\begin{figure*}[t]
\centering
\includegraphics[width=0.95\linewidth]{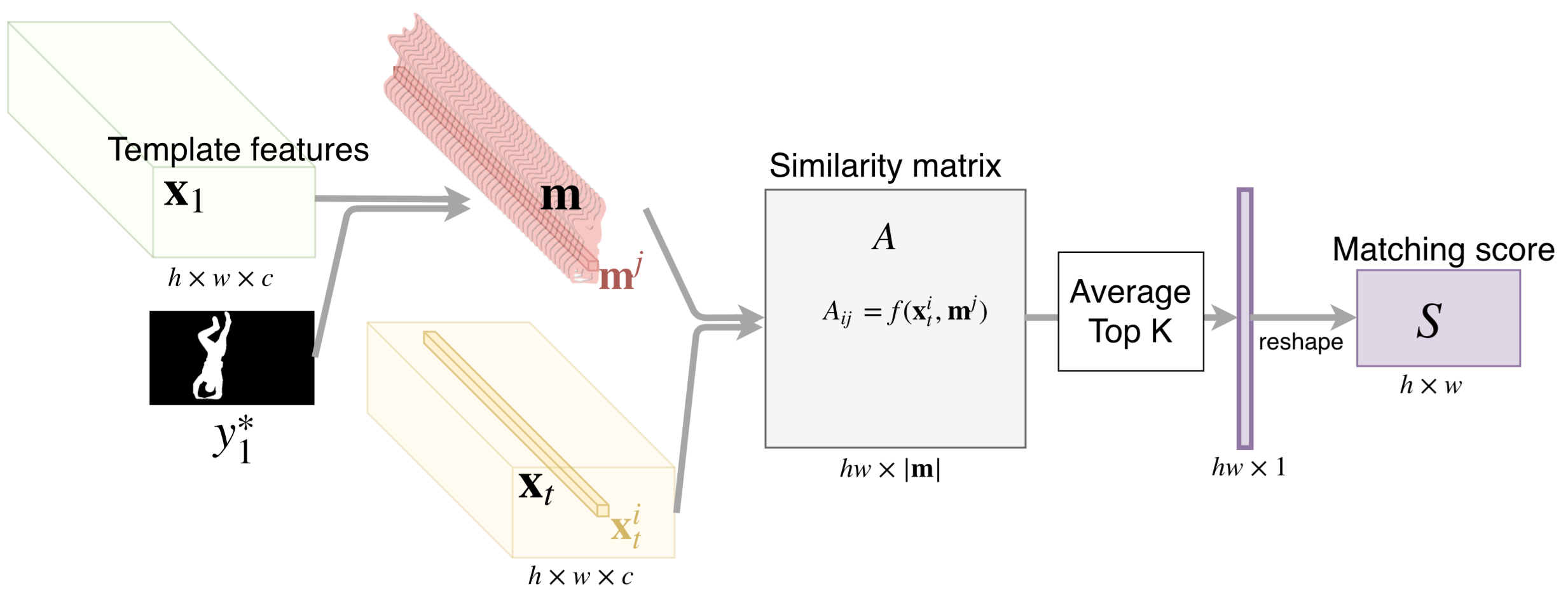}\\
\caption{\tb{Illustration of the proposed soft matching layer.} We first take two sets of features and compute pairwise similarity between all pairs of features. We then produce the final matching score by computing the average of top K similarity scores.
}
\label{fig:SoftMatching}
\end{figure*}
\subsection{Soft matching layer}
\label{sec:SoftMatching}
A schematic illustrating the details of the proposed soft matching layer is given in \figref{fig:SoftMatching}. The developed soft matching layer, $\operatorname{SML}(\bx_t, \bm)$, takes two sets of features as inputs, \ie, $\bx_t$ and $\bm$ ($\bm$ refers to either $\bm_F$ or $\bm_B$) and computes a matching score matrix $S_t\in\mathbb{R}^{h\times w}$ which measures the compatibility of the frame $I_t$ (represented by its features $\bx_t$) with either foreground ($\bm_F$) or background ($\bm_B$) pixels of the template $I_1$ for every pixel $i\in\{1, \ldots, hw\}$. The entry $S_t^i$ represents the similarity of the feature at pixel location $i$ with respect to a subset of features in $\bm$.

More formally, our developed soft matching layer first computes the pairwise similarity score matrix $A\in[-1,1]^{(hw)\times|\bm|}$ where the $ij$-th entry of $A$ is calculated via
$$
A_{ij} = f(\bx_t^i, \bm^j).
$$
Hereby, $f$ is a scoring function measuring the similarity between the two feature vectors $\bx_t^i$ and $\bm^j$. We use the cosine similarity, \ie, $f(\bx_t^i, \bm^j) = \frac{\bx^j_t\cdot \bm^j}{\|\bx_t^j\|\|\bm^j\|}$, but any other distance metric is equally applicable once adequately normalized. 

Given the similarity score matrix $A$, we compute the matching score matrix $S_t$ of size $h\times w$, respectively its $i$-th entry ($i\in\{1, \ldots, hw\}$) via
$$S_t^i=\frac{1}{K}\sum\limits_{j\in \operatorname{Top}(A_i, K)} A_{ij},$$
where the set $\operatorname{Top}(A_i, K)$ contains the indices with the top $K$ similarity scores in the $i$-th row of the similarity score matrix $A$. $K$ is set to $20$ in all our experiments.

Intuitively, we use the average similarity of the top $K$ matches because we assume a pixel to match to a number of pixels in a region as opposed to only one pixel, which will be too noisy, or to all pixels, which will be too strict in general as the foreground or background may be rather diverse. Consequently, we expect a particular pixel to match to one of the foreground or background regions rather than requiring a pixel only to match locally or to all regions. 
Again, an illustration of the soft matching layer, $\operatorname{SML}(\bx_t, \bm)$, is presented in~\figref{fig:SoftMatching}.

\subsection{Outlier removal and online update}
\label{sec:OnlineUpdate}

\setlength{\figwidth}{0.245\textwidth}
\begin{figure*}[t]
\begin{center}
	\begin{subfigure}[b]{\figwidth}
		\includegraphics[width=\linewidth]{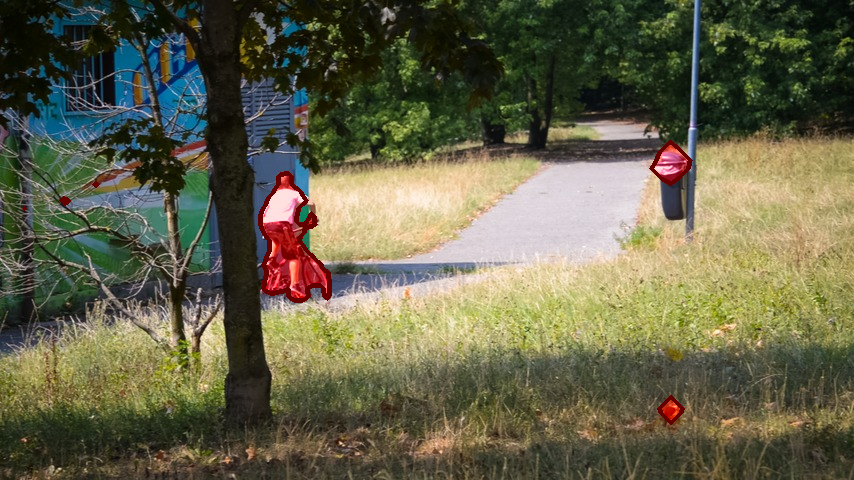} \\
		\centering (a) FG pred. $y_{t,\mathrm{init}}$
	\end{subfigure}\hfill
	\begin{subfigure}[b]{\figwidth}
		\includegraphics[width=\linewidth]{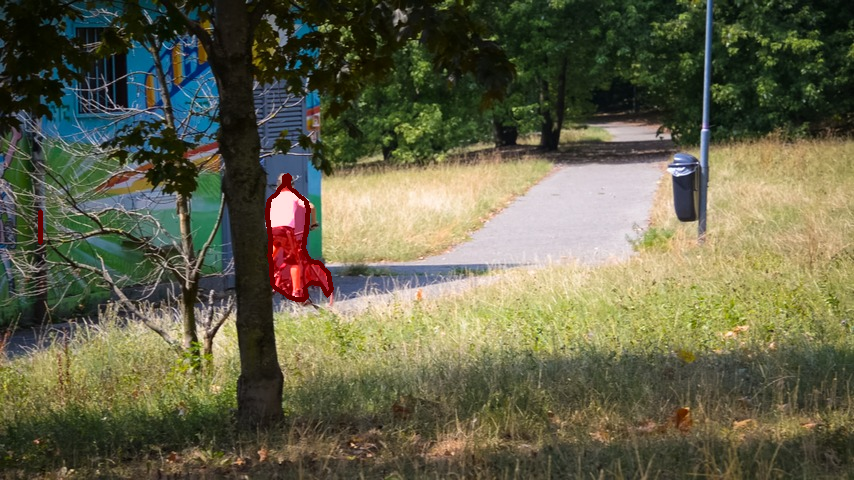} \\
		\centering (b) FG pred. $y_{t-1}$
	\end{subfigure}\hfill
	\begin{subfigure}[b]{\figwidth}
		\includegraphics[width=\linewidth]{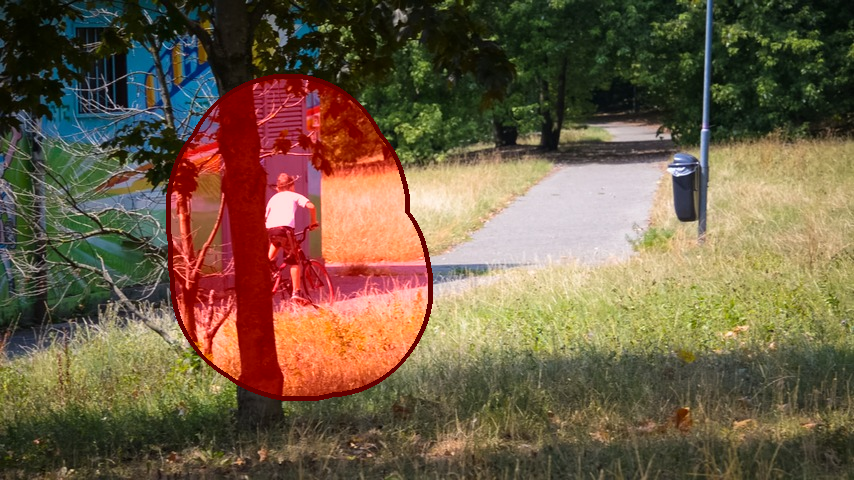} \\
		\centering (c) Extruded pred. $\hat{y}_{t-1}$
	\end{subfigure}\hfill
	\begin{subfigure}[b]{\figwidth}
		\includegraphics[width=\linewidth]{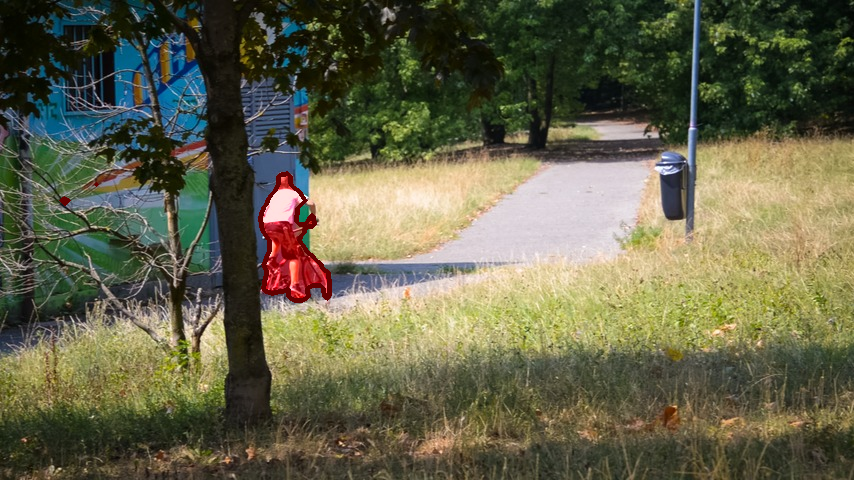} \\
		 \centering (d) Output pred. $y_{t}$
	\end{subfigure}\hfill
\end{center}
\caption{\tb{Example of the proposed outlier removal process.} We first extrude the prediction from the previous frame (b) to obtain an extruded prediction (c). We then produce the prediction at the current frame by finding the intersection between (a) and (c).
}
\label{fig:Online}
\end{figure*}

\para{\bf Outlier removal.} To obtain the final prediction $y_t$ for frame $t\in\{2, \ldots, T\}$ we convert the foreground and background matching score matrices into an initial foreground probability prediction $y_{t,\text{init}}$ via upsampling and via a subsequent weighted softmax operation. Finally, we obtain the prediction $y_t$ by comparing the initial prediction $y_{t,\text{init}}$ with $y_{t-1}$ to remove outliers. More specifically, we first extrude the prediction $y_{t-1}$ of the previous frame to find pixels whose distance to the segmentation is less than a threshold $d_c$. We then compute $y_t$ from $y_{t,\text{init}}$ by removing all initial foreground predictions that don't overlap with the extruded prediction $\hat y_{t-1}$. Note that the hat symbol `$\hat \cdot$' refers to the extrusion operation. This process assumes that the change of the object of interest is bounded from above. In~\figref{fig:Online}, we visualize one example of the current foreground prediction $y_{t,\text{init}}$, previous foreground prediction $y_{t-1}$, the extruded prediction $\hat{y}_{t-1}$, and the final foreground prediction $y_t$.


\para{\bf Online update.} Obviously, we expect the appearance of the object of interest to change over time in a given video. In order to accommodate the appearance change, we repeatedly adjust the foreground and background model during testing. Inspired by~\cite{voigtlaender2017online}, 
we update the foreground and background sets of features, \ie, $\bm_F$ and $\bm_B$, by appending additional features after we predicted the segmentation for each frame. We find the additional features by comparing the initial prediction mask $y_{t,\text{init}}$ for $t\in\{2, \ldots, T\}$ with the extruded prediction $\hat y_{t-1}$ of the previous frame. 

Specifically, we update the background model $\bm_B$ at time $t$ via
$$
\bm_B \leftarrow \bm_B \cup \{\bx_t^i: i \in \bb_t\},
$$
where the index set
$$
\bb_t = \{i: i\in g(y_{t,\text{init}}), i\notin g(\hat{y}_{t-1})\} = \{i: i\in g(y_{t,\text{init}})\setminus g(y_t)\}
$$
subsumes the set of pixels that are predicted as foreground initially, \ie, in $y_{t,\text{init}}$, yet don't belong to the set of foreground pixels in the extruded previous prediction $\hat{y}_{t-1}$. Note that this is equivalent to the set of pixels which are predicted as foreground initially, \ie, $y_{t,\text{init}}$, but are not part of the final prediction $y_t$. Taking~\figref{fig:Online} as an example, $\bb_t$ contains the indices of pixels being foreground in~\figref{fig:Online}(a) but not in ~\figref{fig:Online}(b). 

Intuitively, we find the possible outliers in the current predictions if a pixel is predicted as foreground at time $t$ but does not appear to be foreground or is near to the foreground mask at time $t-1$. 

Beyond adjusting the background model we also update the foreground model $\bm_F$ via
$$
\bm_F \leftarrow \bm_F \cup \{\bx_t^i: i\in g(\breve{y}_t), y_t^i > c, i \notin \bb_t \},
$$
where $g(\breve{y}_t)$ is the set of foreground pixels in the eroded current segmentation prediction $y_t$ and $c$ is a constant threshold. Intuitively, we add the features of pixels that are not only predicted as foreground with high confidence (larger than $c_1$) but are also far from the boundary. In addition, we exclude those pixels in $\bb_t$ to avoid conflicts between the foreground and background features.

Since our method just appends  additional representations to the foreground and background features $\bm_F$ and $\bm_B$, the parameters of the employed network remain fixed, and the online update step is fast. Compared to~\cite{voigtlaender2017online}, where each online update requires fine-tuning the network on the tested images, our approach is more efficient. Note that we designed a careful process to select features which are added in order to avoid the situation that the sizes of $\bm_F$ and $\bm_B$ grow intractably large, which will slow down the computation when computing the matching scores. It is obviously possible to keep track of how frequently features appear in the $\operatorname{Top}-K$ set and remove those that don't contribute much. In practice, we didn't find this to be necessary for the employed datasets. 
\subsection{Instance-level video object segmentation}
\label{sec:Instance}
Next, we explain how the proposed method can be generalized for instance-level video object segmentation, where one or more objects of interest are presented in the first frame of the video. We consider the case where the ground truth segmentation mask contains a single or multiple objects, \ie, $y^\ast_1\in\{1,\ldots,N\}^{H\times W}$, where $N\geq1$. We construct the foreground and background features for every object,~\ie, we find the foreground features $\bm_{F,k}$ and the background features $\bm_{B,k}$ of the object $k\in\{1,\ldots,N\}$, where
$$
\bm_{F,k} = \{\bx_1^i : i \in g(\delta(y_1^\ast=k))\}\quad\quad\text{ and }\quad\quad \bm_{B,k} = \{\bx_1^i : i \notin g(\delta(y_1^\ast=k))\}.
$$
Hereby, $\delta(\cdot): \{1,\ldots,N\}^{H\times W}\rightarrow \{0,1\}^{H\times W}$ is the indicator function which provides a binary output indicating the regions in $y^\ast_1$ that belong to the $k$-th object. We then compute $y_{t,k}$, the foreground probability map of the frame $t$ \wrt the $k$-th object by considering $\bx_t$, $\bm_{F,k}$ and $\bm_{B,k}$ using the soft matching layer described above. After having computed $k$ probability maps, we fuse them to obtain the final output prediction. The prediction $y_t$ is computed by finding the index of the object that has maximum probability $y_{t,k}^i$ among all $k\in\{1,\ldots,N\}$ for all pixels $i$. If for all $k$, $y_{t,k}^i$ is less than a threshold $c_2$, the pixel $i$ will be classified as background.


\section{Experimental Results}
In the following we first provide implementation details before evaluating the proposed approach on a variety of datasets using a variety of metrics.

\subsection{Implementation details, Training and Evaluation}

To obtain the features $\bx$, we found ResNet-101~\cite{HeCVPR16} as the backbone with dilated convolutions~\cite{ChenArxiv16} to perform well. More specifically, we use the representation from the top convolutional layer in the network as $\bx_t$. The feature maps have spatial resolution 8 times smaller than the input image. In the experiments, we set $K=20$, $d_c=100$, $c_1=0.95$ and $c_2=0.4$. We initialized the parameters using the model pretrained on Pascal VOC~\cite{EveringhamIJCV15,HariharanICCV11} for semantic image segmentation. We trained the entire network end-to-end using the Adam optimizer~\cite{kingma2014adam}. We set the initial learning rate to $10^{-5}$ and gradually decreases over time. The weight decay factor is $0.0005$.

To training our matching network, we use any two randomly chosen frames in a video sequence as training pairs. Importantly, the two frames are not required to be consecutive in time which provides an abundance of training data. We augmented the training data by random flipping, cropping and scaling between a factor of 0.5 to 1.5. We use Tensorflow to implement the algorithm.  Training takes around 4 hours for 1000 iterations  on an Nvidia Titan X. 
At test time, a forward pass with an input image of size $480\times 854$ takes around 0.17 seconds. 


\para{\bf Training:} We trained the proposed network using the 30 video sequences available in the DAVIS-16 training set~\cite{PerazziCVPR16} for 1000 iterations and evaluated on the DAVIS-16 validation set. Similarly, we used the 60 sequences in the DAVIS-17 training set~\cite{pont2017DAVIS} for training when testing on the DAVIS-17 validation set. Although the model is trained on DAVIS, we found it to generalize well to other datasets. Therefore, we use the model trained on the DAVIS-17 training set for evaluation on both the JumpCut~\cite{FanTOG15} and the YouTube-Objects~\cite{PrestCVPR12} datasets. 

\para{\bf Evaluation:}
We validate the effectiveness of our method on the DAVIS-16~\cite{PerazziCVPR16} validation, the DAVIS-17~\cite{pont2017DAVIS} validation, the JumpCut~\cite{FanTOG15} and the YouTube-Objects~\cite{PrestCVPR12} datasets. For the YouTube-Objects dataset, we use the subset with groundtruth segmentation masks provided by~\cite{JainECCV2014}, containing 126 video sequences. All of the datasets provide pixel-level groundtruth segmentation. More specifically, binary (foreground-background) ground truth is provided in the DAVIS-16, JumpCut, and YouTube-Objects datasets, while there is instance-level segmentation groundtruth available for the DAVIS-17 dataset. Challenges such as occlusion, fast motion, and appearance change are presented in the four datasets. Thus, these four datasets serve as a good test bed to evaluate different video object segmentation techniques.


\subsection{Evaluation metrics}

\para{\bf Jaccard index (mIoU):} Jaccard index is a common evaluation metric to evaluate the segmentation quality. It is calculated as the intersection over union (IoU) of the predicted and groundtruth masks. We compute the mean of the IoU across all the frames in a sequence and thus also refer to this metric as mIoU.

\para{\bf Contour accuracy (F)~\cite{PerazziCVPR16}:} To measure the quality of the predicted mask, we assess the contour accuracy by computing a bipartite matching between the contour points of the predicted segmentation and the contour points of the groundtruth segmentation. Based on the matching result we calculate the contour accuracy via the F-1 score.

\para{\bf Error rate~\cite{FanTOG15}:} Following the evaluation protocol in~\cite{FanTOG15}, we compute the error rate on the JumpCut dataset. We select key frames $i=\{0,16,...,96\}$ in each sequence and for the $i$-th keyframe, we compute the error in the predicted segmentation of the $i+d$-th frames, given the groundtruth segmentation mask of the $i$-th frame. Intuitively, we measure the transfer (or matching) error of methods with respect to a certain transfer distance $d$. The error is equal to the number of false positive and false negative pixels (the mislabeled pixels) divided by the number of all positive pixels (all foreground pixels) in the predicted segmentation of the $i+d$-th frame. We use $d=16$ in the experiments and compute the average of the errors to obtain the error rate.

\subsection{Quantitative results}

We carefully evaluated the proposed approach and compared the proposed method with a wide variety of video object segmentation methods~\ie, MSK~\cite{PerazziCVPR17}, SFL~\cite{ChenICCV17}, OSVOS~\cite{caelles2016one}, OnAVOS~\cite{voigtlaender2017online}, PLM~\cite{yoon2017pixel}, MaskRNN~\cite{hu2017maskrnn}, Lucid~\cite{khoreva2017lucid}, SEA~\cite{AvinashCVPR14}, HVS~\cite{GrundmannCVPR2010}, JMP~\cite{FanTOG15}, FCP~\cite{PerazziICCV15}, BVS~\cite{MaerkiCVPR16}, OFL~\cite{TsaiCVPR2016}, CTN~\cite{JangCVPR17}, VPN~\cite{JampaniCVPR17}, 
SVC~\cite{WangTIP2017}, JFS~\cite{NagarajaICCV15}, LTV~\cite{OchsPAMI2014}, HBT~\cite{GodecICCV2011}, AFS~\cite{VijayanarasimhanECCV2012}, SCF~\cite{JainECCV2014}, RB~\cite{BaiSIGGRAPH2009} and DA~\cite{ZhongTOG12}. Note that MSK, OSVOS, SFL, OnAVOS, PLM, MaskRNN, Lucid employ fine-tuning during testing.

We present the quantitative results on four datasets: DAVIS-16~\cite{PerazziCVPR16}, YouTube-Objects~\cite{PrestCVPR12}, JumpCut~\cite{FanTOG15} and DAVIS-17~\cite{pont2017DAVIS}. Our method outperforms state-of-the-art methods by $0.4\%$ in mIoU and by 0.71 in error rate on Youtube-Objects and JumpCut datasets, respectively. On DAVIS-16 and DAVIS-17 datasets, our approach performs on par with state-of-the-art techniques while not using fine-tuning. The quantitative results are summarized in \tabref{tab:nofinetuning},~\ref{tab:youtube},~\ref{tab:jumpcut},~\ref{tab:davis17} and \figref{fig:davis16}. The best method is highlighted in bold and the second-best method is underlined. Details are described in the following.

\para{\bf Evaluation on the DAVIS-16 dataset:}
In~\tabref{tab:nofinetuning}, we compare our method with deep net baselines that do not require fine-tuning as well, such as VPN~\cite{JampaniCVPR17} and CTN~\cite{JangCVPR17}. We also compare to OSVOS~\cite{caelles2016one}, MSK~\cite{PerazziCVPR17}, OnAVOS~\cite{voigtlaender2017online} and SFL ~\cite{ChenICCV17}, disabling their fine-tuning step. We use the super-script `$^-$' to denote methods with a disabled fine-tuning step. In~\tabref{tab:nofinetuning}, we report the mean IoU and the average running time per frame for each method tested on the DAVIS-16 dataset. Our method achieves the best mIoU, outperforming the baselines by more than 6\% while running efficiently. Our method without the outlier removal (denoted as OURS-NU in~\tabref{tab:nofinetuning}) runs 2 times faster while achieving competitive performance.

In~\figref{fig:davis16}, we compare our method which does not require fine-tuning with baselines that may or may not need fine-tuning. We report the mIoU vs average computational time per frame in~\figref{fig:davis16}(a) and the contour accuracy vs running time per frame in~\figref{fig:davis16}(b). Note that the average running time per frame also includes the fine-tuning step for those methods requiring fine-tuning. Since the network employed in our method is general enough to learn how to match we observe competitive performance at a fraction of the time required by other techniques. Note that the time axis scaling is logarithmic.

\begin{table}[t]
	\centering
	{\small
		\caption{Comparisons with deep net methods \emph{without} fine-tuning (VPN and CTN) or with fine-tuning step disabled (denoted with $^-$) on DAVIS-16 validation set. OURS-NU: our method without online update and outlier removal.
		}
		\label{tab:nofinetuning}
		\tabcolsep=5pt
		\resizebox{0.9\linewidth}{!}{
\begin{tabular}{lcccccccc}
	\toprule
		&OURS            & OURS-NU & OSVOS$^-$ & MSK$^-$  & OnAVOS$^-$ & SFL$^-$  & VPN   & CTN    \\
	\midrule
	mIoU             & \bf 0.810            & {\ul 0.792}  & 0.525 & 0.699   & 0.736 & 0.674 & 0.702 & 0.735 \\
	Speed (s) & 0.32              & 0.17   & \bf 0.12  & {\ul 0.15}    & 3.55  & 0.3   & 0.63  & 29.95\\
	\bottomrule
\end{tabular}
		}
	}
\end{table}


\setlength{\figwidth}{0.495\textwidth}
\begin{figure*}[t]
\begin{center}
	   \hfill
		\begin{subfigure}[b]{\figwidth}
		\includegraphics[width=\linewidth,height=0.8\linewidth]{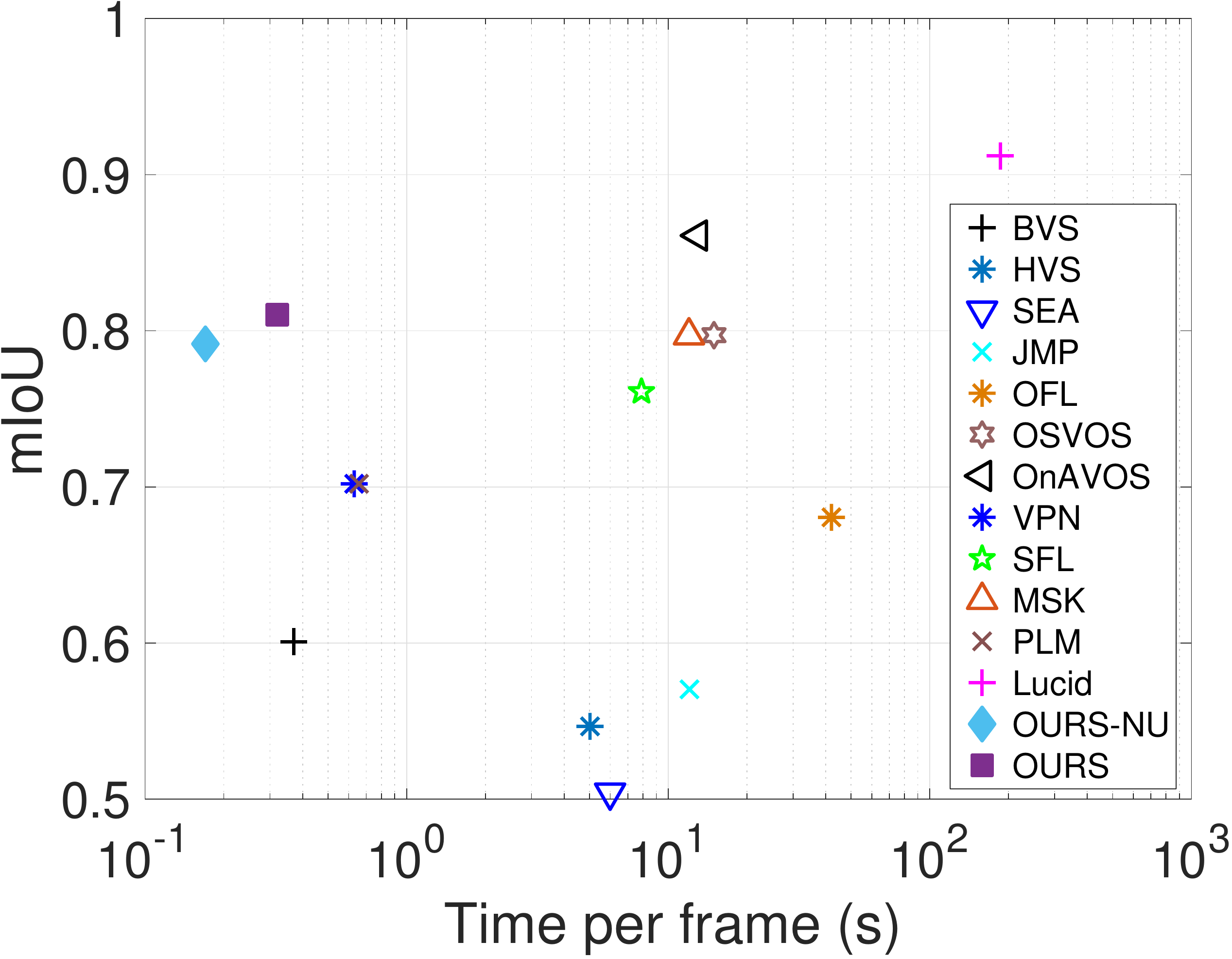} \\
		\centering (a) mIoU \vs speed
	\end{subfigure}\hfill
	\begin{subfigure}[b]{\figwidth}
		\includegraphics[width=\linewidth,height=0.8\linewidth]{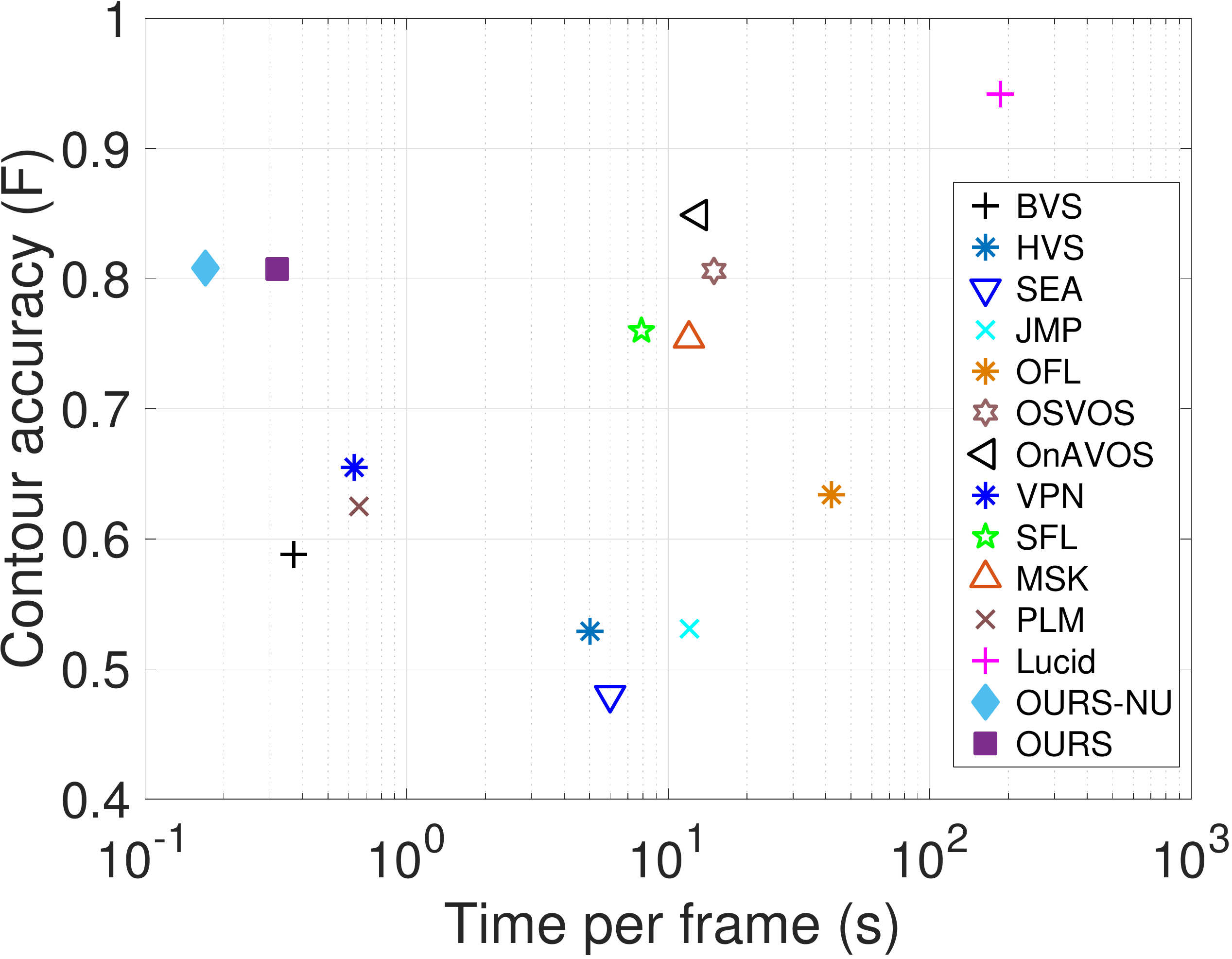} \\
		\centering (b) F \vs speed
	\end{subfigure}\hfill
\end{center}
\caption{\tb{Performance comparison on the DAVIS-16 validation set.} The x axis denotes the average running time per frame in seconds (log scale) and the y axis is  (a) mIoU (Jaccard index) and (b) F score (contour accuracy). }
\label{fig:davis16}
\end{figure*}

\para{\bf Evaluation on the YouTube-Objects dataset:}
We present the evaluation results on the YouTube-Objects dataset~\cite{PrestCVPR12,JainECCV2014} in~\tabref{tab:youtube}. Our method outperforms the baselines despite the fact that our network is not fine-tuned, but other baselines such as OnAVOS and MSK and OSVOS are. Thus, our method is more favorable both in terms of computational time and in terms of accuracy.

\begin{table}[t]
	\centering
	{\footnotesize
		\caption{Evaluation on the Youtube-Object dataset~\cite{PrestCVPR12,JainECCV2014} using Jaccard index (mIoU).}
		\label{tab:youtube}
		\tabcolsep=3pt
		\resizebox{\linewidth}{!}{
\begin{tabular}{lcccccccccccc}
	\toprule
	Sequence  & OURS  & OnAVOS & MSK   & OSVOS & OFL   & JFS   & BVS   & SCF   & AFS   & FST   & HBT   & LTV   \\
	\midrule
	Fine-tuned?& - & Yes & Yes & Yes & - & - & - & - & - & - & - & -\\
	\midrule
	Aeroplane & 0.880 & 0.902  & 0.816 & 0.882 & 0.899 & 0.89  & 0.868 & 0.863 & 0.799 & 0.709 & 0.736 & 0.137 \\
	Bird      & 0.873 & 0.879  & 0.829 & 0.857 & 0.842 & 0.816 & 0.809 & 0.81  & 0.784 & 0.706 & 0.561 & 0.122 \\
	Boat      & 0.805 & 0.816  & 0.747 & 0.775 & 0.74  & 0.742 & 0.651 & 0.686 & 0.601 & 0.425 & 0.578 & 0.108 \\
	Car       & 0.779 & 0.738  & 0.670 & 0.796 & 0.809 & 0.709 & 0.687 & 0.694 & 0.644 & 0.652 & 0.339 & 0.237 \\
	Cat       & 0.788 & 0.759  & 0.696 & 0.708 & 0.683 & 0.677 & 0.559 & 0.589 & 0.504 & 0.521 & 0.305 & 0.186 \\
	Cow       & 0.771 & 0.787  & 0.750 & 0.778 & 0.798 & 0.791 & 0.699 & 0.686 & 0.657 & 0.445 & 0.418 & 0.163 \\
	Dog       & 0.803 & 0.809  & 0.752 & 0.813 & 0.766 & 0.703 & 0.685 & 0.618 & 0.542 & 0.653 & 0.368 & 0.18  \\
	Horse     & 0.688 & 0.742  & 0.649 & 0.728 & 0.726 & 0.678 & 0.589 & 0.54  & 0.508 & 0.535 & 0.443 & 0.115 \\
	Motorbike & 0.774 & 0.663  & 0.498 & 0.735 & 0.737 & 0.615 & 0.605 & 0.609 & 0.583 & 0.442 & 0.489 & 0.106 \\
	Train     & 0.811 & 0.838  &  0.777 & 0.757 & 0.763 & 0.782 & 0.652 & 0.663 & 0.624 & 0.296 & 0.392 & 0.196 \\
	\midrule
	Average   & \bf{ 0.797} & {\ul 0.793}  & 0.718 & 0.783 & 0.776 & 0.74  & 0.68  & 0.676 & 0.625 & 0.538 & 0.463 & 0.155\\
	\bottomrule
\end{tabular}	
}
}
\end{table}

\para{\bf Evaluation on the JumpCut dataset:}
We present the evaluation results on the JumpCut dataset~\cite{FanTOG15} in~\tabref{tab:jumpcut}.  We follow the evaluation in~\cite{FanTOG15} and compute the error rates of different methods. The transfer distance $d$ is equal to $16$. In this experiment we don't apply the outlier removal described in \secref{sec:OnlineUpdate} to restrict mask transfer between non-successive frames. 
Again, our method outperforms the baselines on this dataset with an average error rate that is 0.34 lower than the best competing baseline SVC~\cite{WangTIP2017}.
\begin{table}[t]
	\centering
	{\small
		\caption{Error rates on the JumpCut dataset~\cite{FanTOG15}. The transfer distance $d$ is $16$.
		}
		\label{tab:jumpcut}
		\tabcolsep=3pt
		\resizebox{\linewidth}{!}{
\begin{tabular}{llrrrrrrrllrrrrrrr}
	\toprule
	&               & RB  & DA  & SEA   & JMP   & SVC   & PLM   & OURS &         &            & RB  & DA  & SEA   & JMP   & SVC   & PLM   & OURS \\ \hline

	&Fine-tuned? &- & - & - & - & -& Yes & - &  &  & - & - & - & - & - & Yes & -\\
	\midrule
	ANIMAL & bear          & 4.58  & 4.48  & 4.21  & 4     & {\bf 2.11}  & {\ul 3.45}  & 5.14 & SNAPCUT & animation  & 11.9  & 6.38  & 6.78  & {\ul 4.55}  & \bf 3.35  & 5.86  & 6.15 \\
	& giraffe       & 22    & 11.2  & 17.4  & \bf 7.4   & {\ul 9.67}  & 17.4  & 11.96 &         & fish       & 51.8  & 21.7  & 25.7  & 17.5  & {\ul 7.67}  & \bf 7.42  & 12.21 \\
	& goat          & 13.1  & 13.3  & 8.22  & \bf 4.14  & 4.97  & 15.2  & {\ul 4.73} &         & horse      & 8.39  & 45.1  & 37.8  & {\ul 6.8}   & \bf 4.84  & 7.94  &  8.25\\ \cline{11-18} 
	& pig           & 9.22  & 9.85  & 10.3  & {\ul 3.43}  & \bf 3.24  & 5.15  & 5.12 &         & Avg.       & 24.03 & 24.39 & 23.43 & 9.62  & \bf 5.29  & {\ul 7.07}  & 8.87 \\ \cline{2-18} 
	& Avg.          & 12.23 & 9.71  & 10.03 &\bf 4.74  & {\ul 5.00}  & 10.30 & 6.74 & FAST    & bball      & 18.4  & 8.47  & 8.89  & \bf 3.9   & {\ul 4.16}  & 8.04  & 6.19 \\ \cline{1-9}
	HUMAN  & couple        & 17.5  & 16    & 23.4  & \bf 5.13  & {\ul 8.49}  & 9.14  & 11.77 &         & cheetah    & 31.5  & 16.6  & 7.68  & 8.16  & \bf 7.1   & 11.8  &  {\ul 7.61} \\
	& park          & 11.8  & 6.54  & 6.91  & {\ul 5.39}  & \bf 5.33  & 10.2  & 11.42 &         & dance      & 56.1  & 50.8  & 43    & 18.7  & 26.5  & \bf 14.7  & {\ul 17.31} \\
	& station       & 8.85  & 20.9  & 21.3  & 9.01  &{\ul 8.42}  & \bf 4.68  & 9.98 &         & hiphop     & 67.5  & 51.1  & 33.7  & 14.2  & 21.9  & {\ul 13.6}  & \bf 10.49 \\ \cline{2-9}
	& Avg.          & 12.72 & 14.48 & 17.20 & \bf 6.51  & {\ul 7.41}  & 8.01  & 11.06 &         & kongfu     & 40.2  & 40.8  & 17.9  & 8     & \bf 3.77  & 6.25  & {\ul 4.05} \\ \cline{1-9}
	STATIC & car           & {\bf 1.76}  & 5.93  & 5.08  & 2.26  & 2.57  & 2.18  & {\ul 1.86} &         & skater     & 38.7  & 40.8  & 29.6  & 22.8  & 21.4  &{\bf 12.6}  & {\ul 13.57} \\
	& cup           & 5.45  & 12.9  & 9.31  & \bf 2.15  & {\ul 2.4}   & 6.04  & 5.38 &         & supertramp & 129   & 60.5  & 57.4  & 42.9  & 27.1  & \bf 20.7  & {\ul 22.12} \\
	& pot           & {\ul 2.43}  & 5.03  & 2.98  & 2.95  & \bf 1.79  & 2.66  & 5.55 &         & tricking   & 79.4  & 70.9  & 35.8  & 21.3  & 21.2  & {\ul 15.7}  &  \bf 8.32 \\ \cline{11-18} 
	& toy           & \bf 1.28  & 3.19  & 2.16  & {\ul 1.3}   & 1.49  & 2.25  & 2.81 &         & Avg.       & 57.60 & 42.50 & 29.25 & 17.50 & 16.64 & {\ul 12.92} & \bf 11.21 \\ \cline{2-9}
	& Avg.          & 2.73  & 6.76  & 4.88  & {\ul 2.17}  & \bf 2.06  & 3.28  & 3.90 &         &            &       &       &       &       &       &       &      \\ 
	\midrule[1pt]
	Average & & 28.68  & 23.75 & 18.89 & 9.82 & {\ul 9.07} & 9.23 & \bf 8.73 &         &            &       &       &       &       &       &       &     \\
	\bottomrule
\end{tabular}
}
}
\end{table}

\para{\bf Evaluation on the DAVIS-17 dataset:}
\begin{table}[t]
	\centering
	{\footnotesize
		
		\caption{Evaluation on the DAVIS-17 validation set.
		}
		\label{tab:davis17}
		\tabcolsep=2pt
\resizebox{\linewidth}{!}{
\begin{tabular}{lcccccccccc}
	\toprule
	& OURS            & OFL   & OSVOS$^-$ & OnAVOS$^-$ & MaskRNN$^-$ & OSVOS & OnAVOS & MaskRNN & OnAVOS$^+$ & OURS-FT   \\
	\midrule
	Fine-tuned? & -&-&-&-&-&Yes&Yes&Yes&Yes&Yes\\
	\midrule
	mIoU             & 0.565 & 0.549  & 0.366   & 0.395    & 0.455 & 0.521  & 0.610    & 0.605   & \bf 0.645             & {\ul 0.614} \\
	Speed (s) & {\ul 0.35}  & 130    & \bf 0.13    & 3.78     & 0.6   & 5    & 13      & 9     & 30                & 2.62 \\
	\bottomrule
\end{tabular}
}
}
\end{table}
We show the experiments on instance-level video object segmentation using the DAVIS-17 validation set. The results are shown in~\tabref{tab:davis17}. Our method performs reasonably well when compared to methods without finetuning,~\ie, OSVOS$^-$, OnAVOS$^-$, MaskRNN$^-$ and OFL. We further finetune our method (denoted as OURS-FT), and the performance is competitive among the baselines while the computational time is much faster. Note that OnAVOS$^+$~\cite{DAVIS2017_5th} in~\tabref{tab:davis17} is OnAVOS with upsampling layers on top and model ensembles.


\setlength{\figwidth}{0.35\textwidth}
\begin{figure*}[t]
\begin{center}
	\begin{subfigure}[b]{\figwidth}
		\includegraphics[width=\linewidth,height=0.8\linewidth]{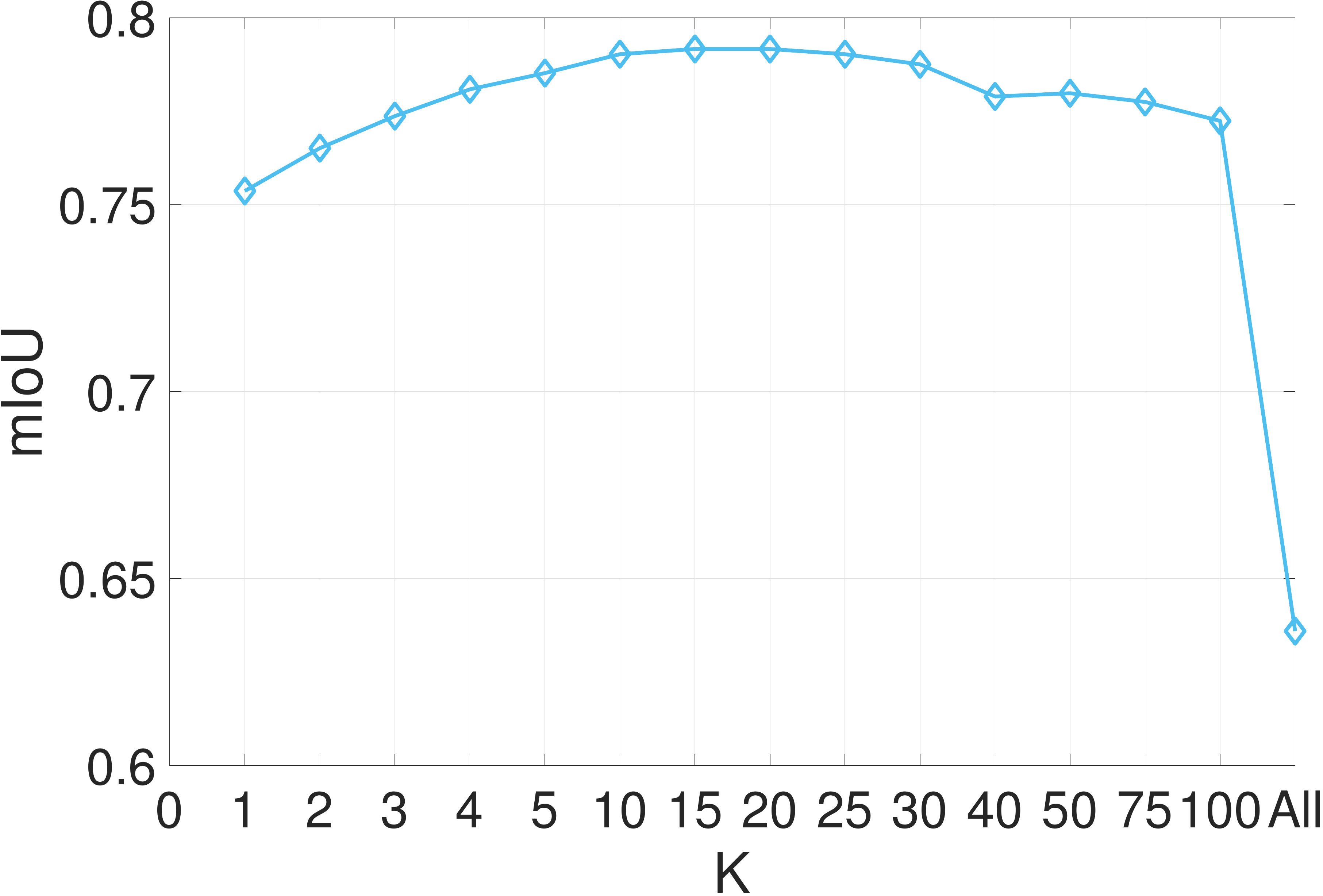} \\ 
		\centering (a) Effect of K in Top K 
	\end{subfigure} \quad\quad
	\begin{subfigure}[b]{\figwidth}
		\includegraphics[width=\linewidth,height=0.8\linewidth]{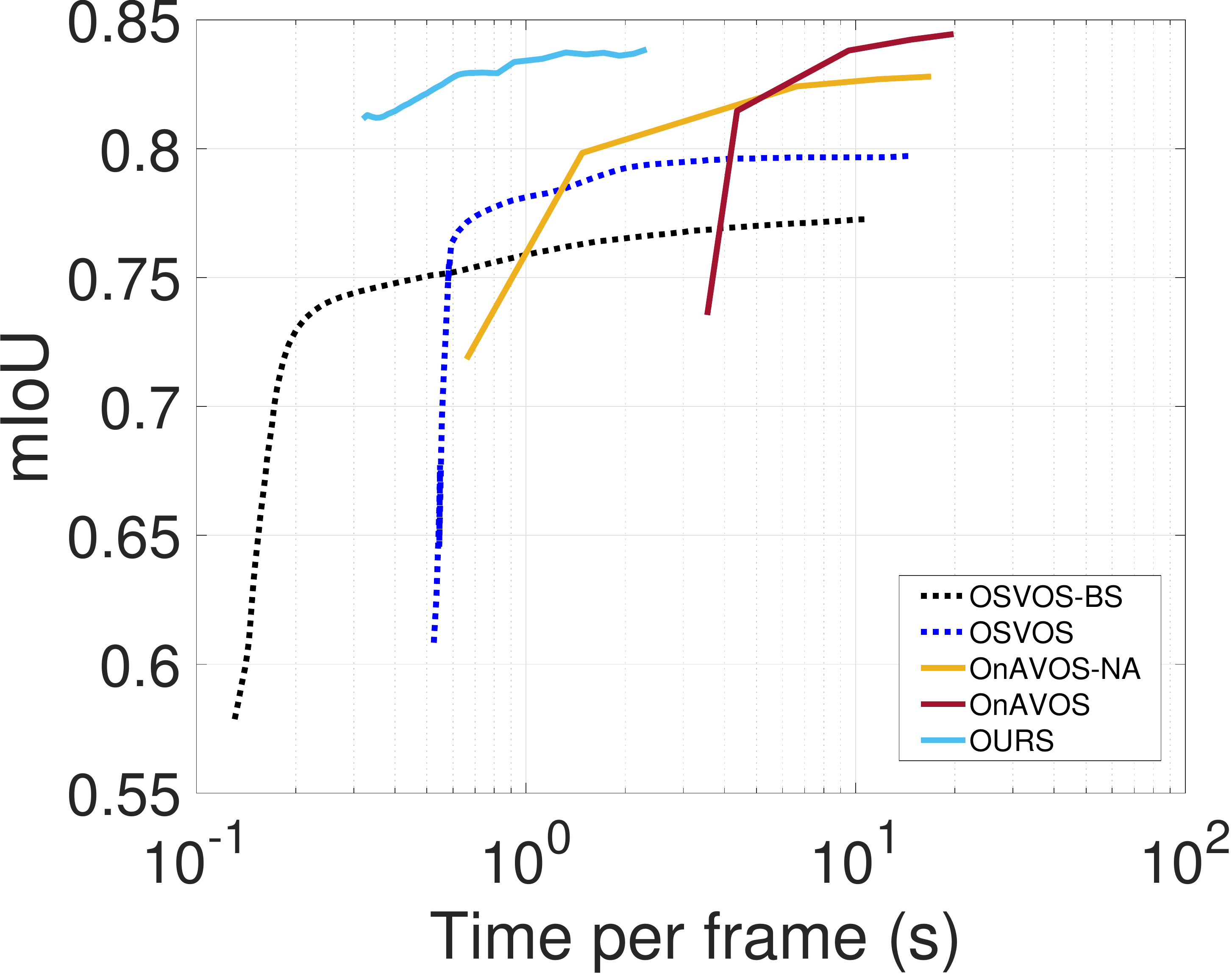} \\
		\centering (b) Effect of fine-tuning
	\end{subfigure}\hfill
\end{center}
\caption{\tb{Sensitivity analysis and finetuning.} (a) The effect of K when computing the Top K similarity scores  in the soft matching layer. (b) The effect of fine-tuning of our approach compared with other baselines. Both results are shown using the DAVIS-16 validation dataset.}
\label{fig:ablation}

\end{figure*}

\subsection{Ablation study}

We study the important components of the proposed method. Subsequently, we discuss the effect of outlier removal and online update, the effect of $K$, the effect of foreground and background matching, the effect of fine-tuning and the memory consumption of the proposed approach.

\begin{table}[t]
\centering
\caption{Ablation study of the three modules in our approach: (1)  outlier removal, (2) online background update, and (3) online foreground update, assessed on the DAVIS-16 validation set.}
\label{tab:ablation}
		\tabcolsep=7pt
\begin{tabular}{cccc}
	\toprule
Outlier removal & BG update & FG update & mIoU  \\
\midrule
-               		& -         & -         & 0.792 \\
$\checkmark$   & -         & -         & 0.805 \\
$\checkmark$   & $\checkmark$ & -         & 0.809 \\
$\checkmark$   & $\checkmark$ & $\checkmark$ & 0.810 \\
\bottomrule
\end{tabular}
\end{table}

\para{\bf Effect of $K$:} We study the effect of $K$ in the proposed soft matching layer where we compute the average similarity scores of top $K$ matchings. We present the performance on DAVIS-16 with different settings of K in~\figref{fig:ablation}~(a). We varied $K$ to be between $1$ and $100$. The performance when $K$ is equal to $1$ (`hard matching') is $0.753$ while the performance increases when $K$ is larger than $1$ (`soft matching') until $K$ is equal to $20$. When $K$ is larger than $20$, the performance keeps decreasing and the performance of computing the average similarity scores among all matchings is $0.636$. Intuitively, a point is a good match to a region if the feature of the point is similar to a reasonable amount of pixels in that region, which motivates the proposed  soft matching layer.

\para{\bf Outlier removal and online update:} 
In~\tabref{tab:ablation}, we study the effects of outlier removal, online background feature update and foreground feature update. We found that our method with neither outlier removal nor online update performs competitively, achieving $0.792$ on DAVIS-16. Removing of outliers improves the performance by $0.013$. If we incorporate  the online background feature update, the performance improves by $0.004$ and having the foreground feature updated as well further improves the performance, achieving $0.810$ in mIoU on the DAVIS-16 dataset.


\setlength{\figwidth}{0.25\textwidth}
\begin{figure}[t]
	\begin{center}
		\begin{subfigure}[b]{0.98\figwidth}%
			\includegraphics[width=\linewidth]{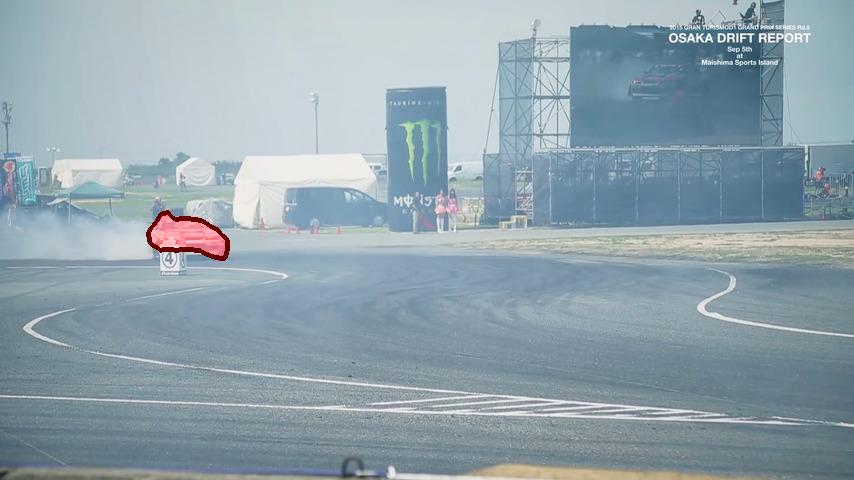} \\
			\includegraphics[width=\linewidth]{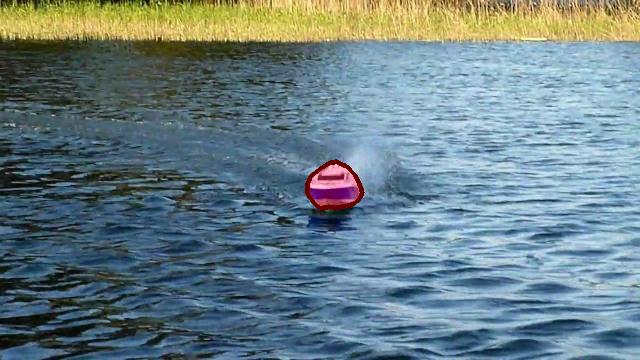} \\
			\includegraphics[width=\linewidth]{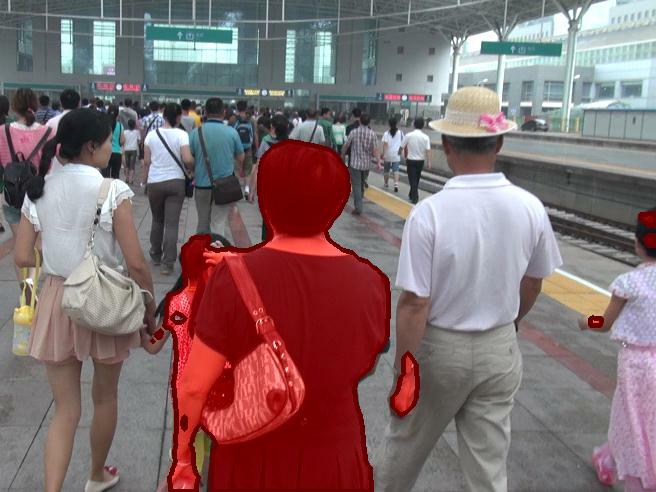} \\
			\includegraphics[width=\linewidth]{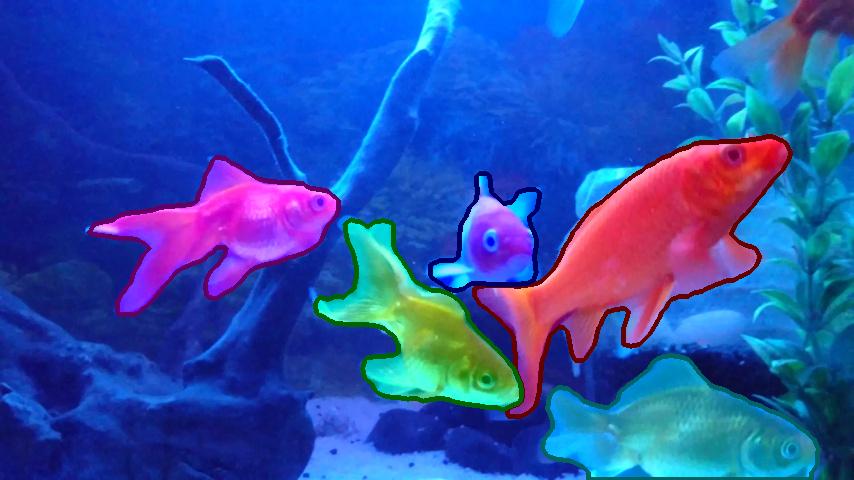} \\
		\end{subfigure}\hfill
		\begin{subfigure}[b]{0.98\figwidth}%
			\includegraphics[width=\linewidth]{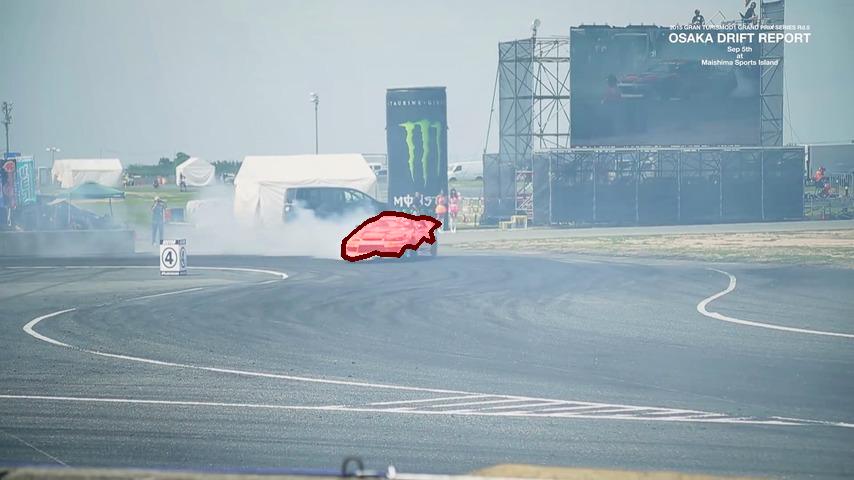} \\
			\includegraphics[width=\linewidth]{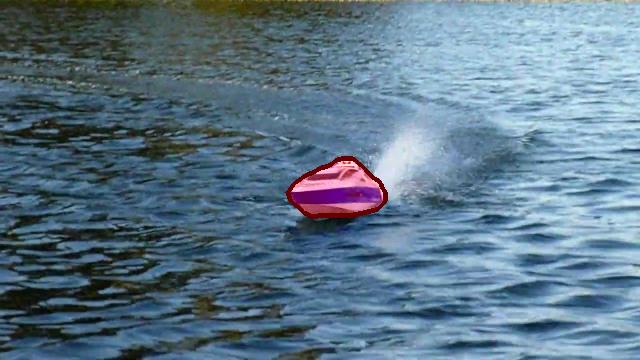} \\
			\includegraphics[width=\linewidth]{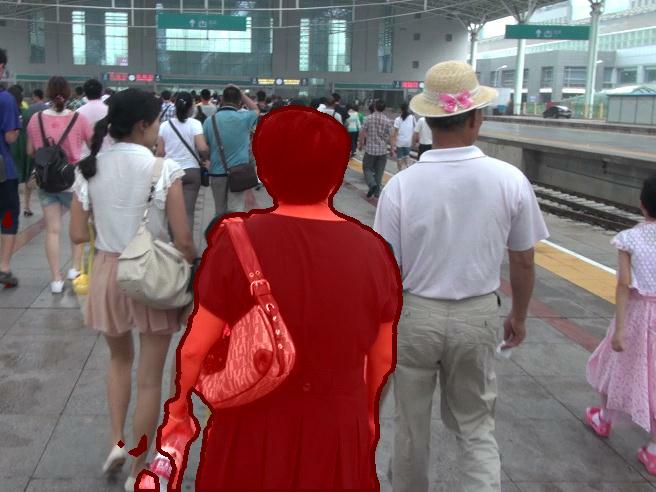} \\
			\includegraphics[width=\linewidth]{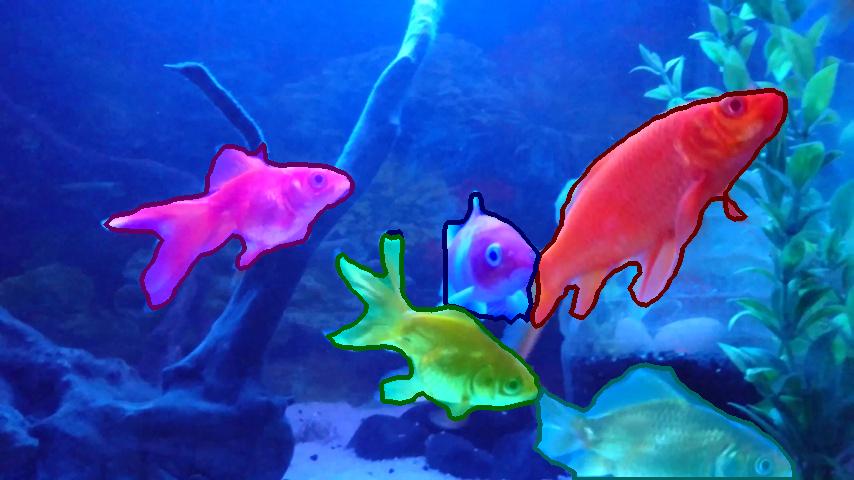} \\
		\end{subfigure}\hfill
		\begin{subfigure}[b]{0.98\figwidth}%
		\includegraphics[width=\linewidth]{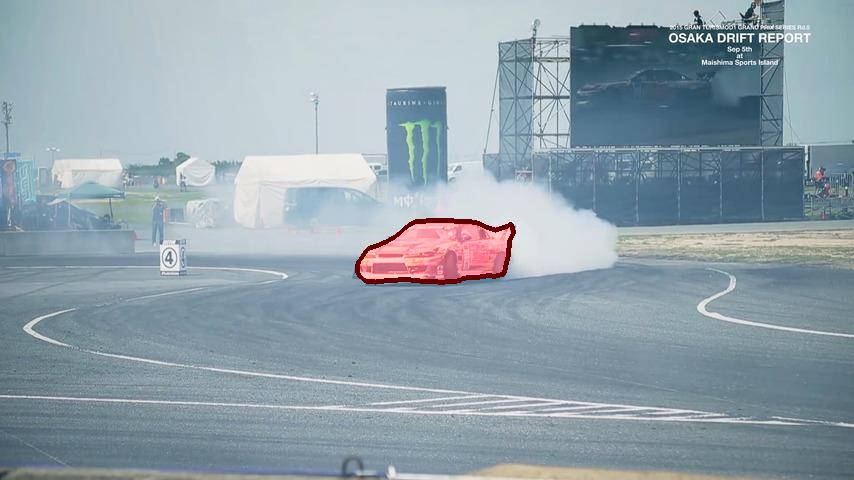} \\
		\includegraphics[width=\linewidth]{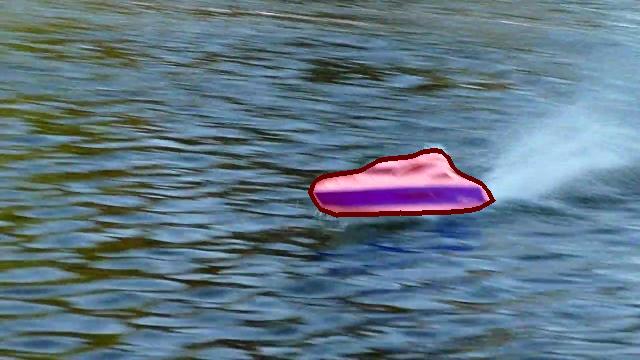} \\
		\includegraphics[width=\linewidth]{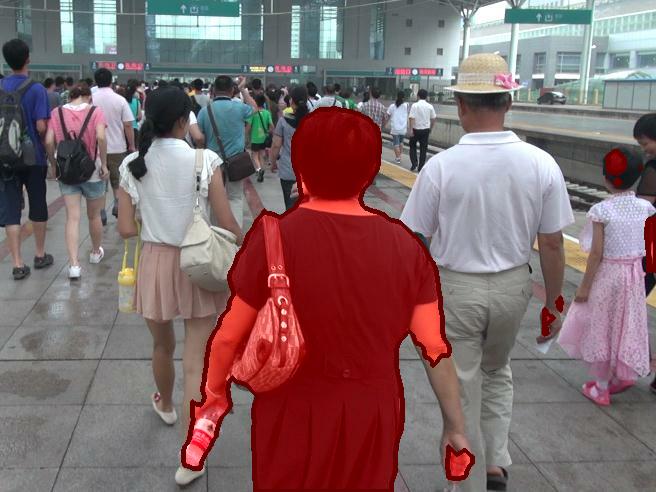} \\
		\includegraphics[width=\linewidth]{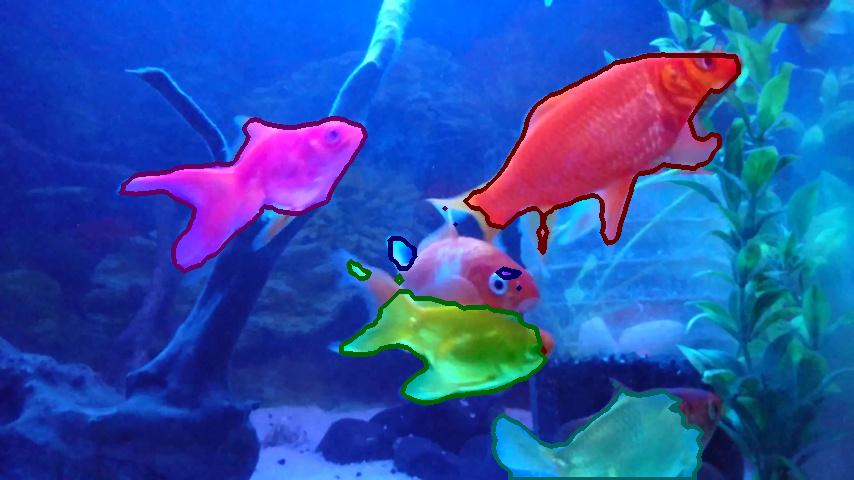} \\
		\end{subfigure}\hfill
		\begin{subfigure}[b]{0.98\figwidth}%
			\includegraphics[width=\linewidth]{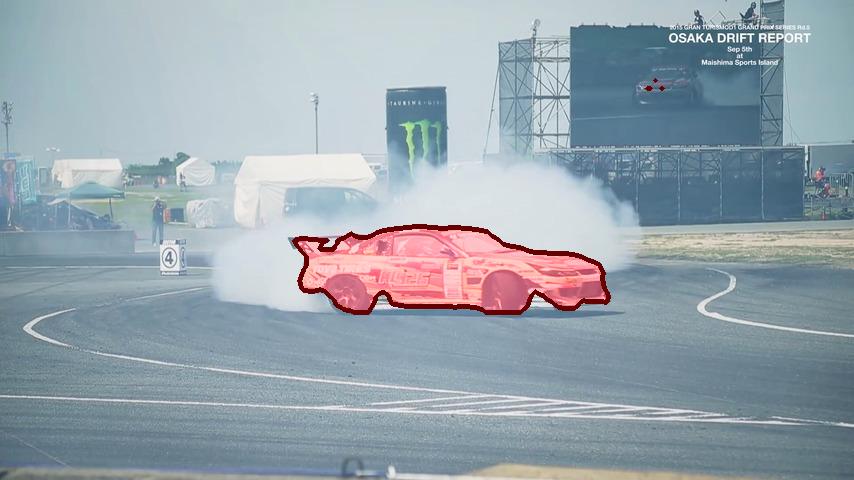} \\
			\includegraphics[width=\linewidth]{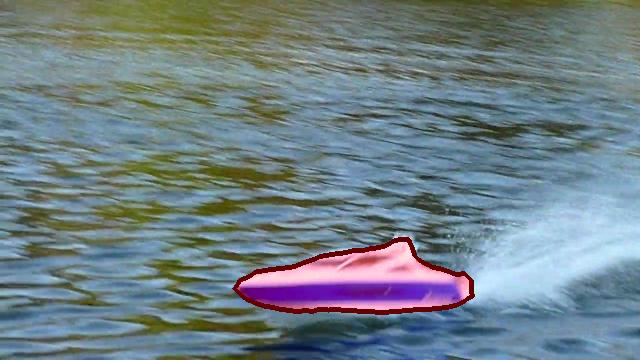} \\
			\includegraphics[width=\linewidth]{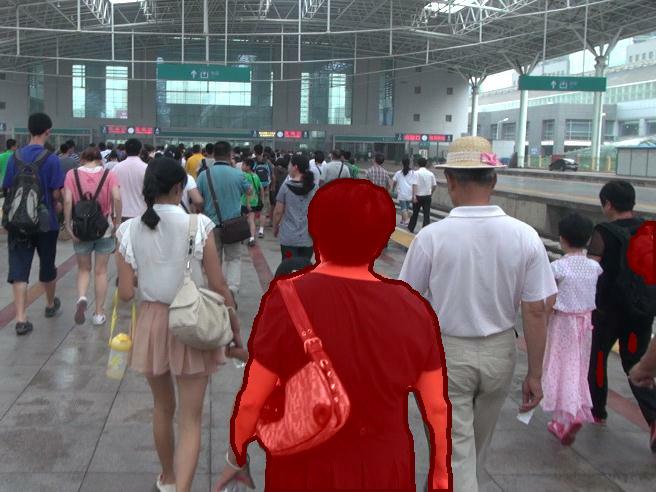} \\
			\includegraphics[width=\linewidth]{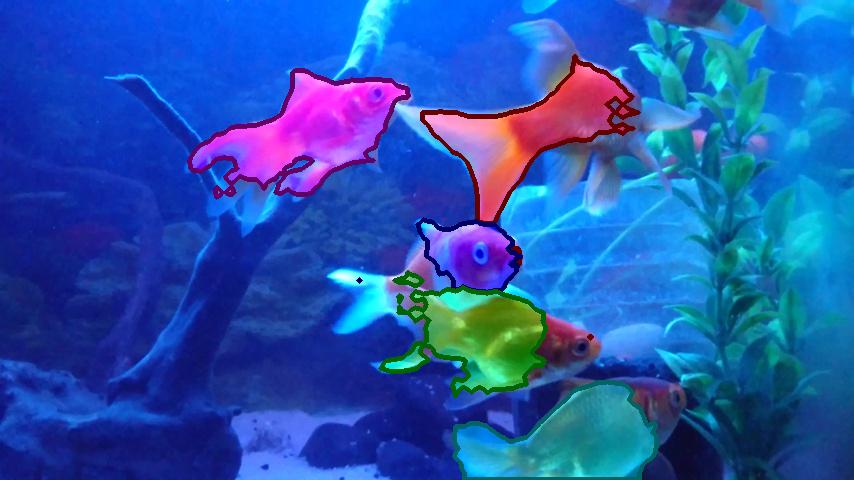} \\
		\end{subfigure}
	\end{center}
	\caption{\tb{Visual results of our approach}. Testing videos are from DAVIS-16 (1\nd{st} row), Youtube-Objects (2\nd{nd} row), JumpCut (3\nd{rd} row),  and DAVIS-17  datasets (4\nd{th} row).
}
	\label{fig:qual}
%
\setlength{\figwidth}{0.25\textwidth}
%
	\begin{center}
		\begin{subfigure}[b]{0.98\figwidth}%
			\includegraphics[width=\linewidth]{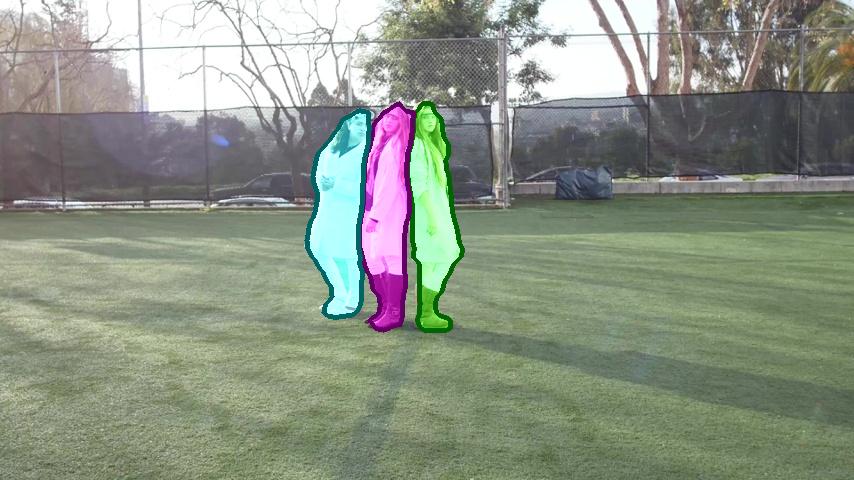} \\
		\end{subfigure}
		\begin{subfigure}[b]{0.98\figwidth}%
			\includegraphics[width=\linewidth]{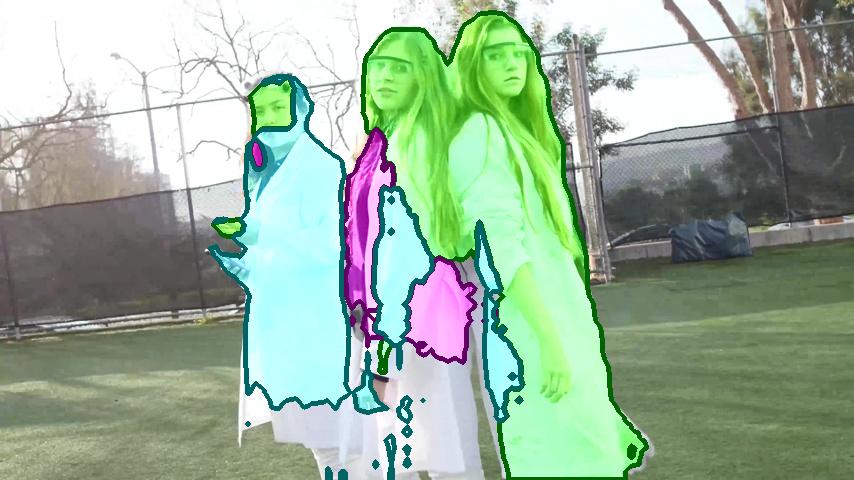} \\
		\end{subfigure}\hfill
		\begin{subfigure}[b]{0.98\figwidth}%
			\includegraphics[width=\linewidth]{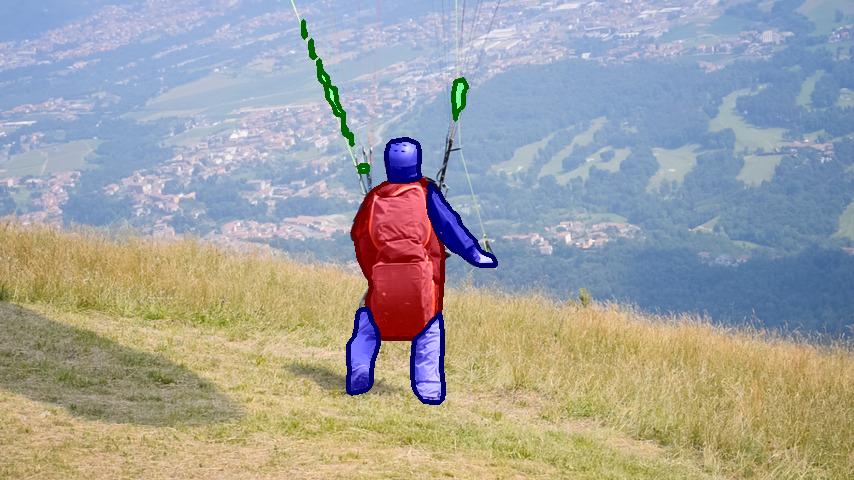} \\
		\end{subfigure}
		\begin{subfigure}[b]{0.98\figwidth}%
			\includegraphics[width=\linewidth]{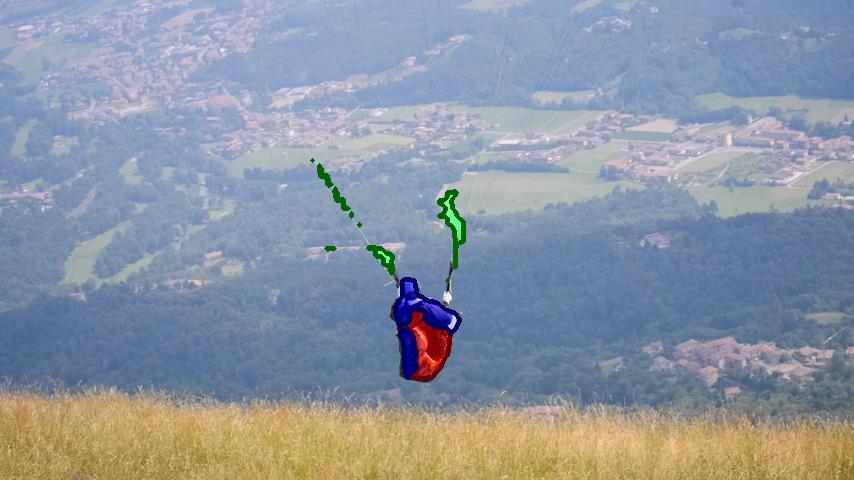} \\
		\end{subfigure}
	\end{center}
	\caption{\tb{Failure cases of our approach.} For each case, we show the results of our approach at the beginning and toward the end of the video sequence.}
	\label{fig:failure}
\end{figure}

\para{\bf Matching foreground and background:}
As shown in~\figref{fig:Overview}, we match the input image with not only the foreground region but also the background region in the template and thus we have two soft matching layers for computing the foreground similarity and the background similarity. We found that having both foreground and background models is important for good performance. Specifically, the performance of matching only the foreground,~\ie, only having one soft matching layer to compute foreground similarity, is only $0.527$ in mIoU on DAVIS-16 while having both foreground and background similarity computed achieves $0.792$. 

\para{\bf Online fine-tuning:} We would like to point out that the network in our method can be fune-tuned during testing when observing the groundtruth mask of the first frame. We show the trade-off between fine-tuning time and performance on DAVIS-16 in~\figref{fig:ablation}~(b). Specifically, we show the average running time per frame taking the fine-tuning step into account, and  compare with OSVOS, OSVOS-BS (OSVOS without the post-processing step), OnAVOS and OnAVOS-NA (OnAVOS without test time augmentation). We report the results of OnAVOS and OnAVOS-NA without a CRF as post-processing. Note that the time axis scaling is again logarithmic. The bottom left  point of each curve denotes performance without fine-tuning. Clearly, the performance of our approach outperforms other baselines if fine-tuning is prohibited. After fine-tuning, our method can be further improved and still runs efficiently, taking 2.5 seconds per frame while other baselines require more than 10 seconds to achieve their peak performance. Note that we don't have any post-processing step to refine the segmentation mask in our method while still achieving competitive results. 


\vspace*{-0.15cm}
\subsection{Qualitative results}

In~\figref{fig:qual}, we show  visual results of our method on DAVIS-16 (1st row), Youtube-Objects (2nd row), JumpCut (3rd row),  and DAVIS-17 datasets (4th row). We observe our method can accurately segment the foreground objects with challenges such as fast motion, cluttered background and appearance change. We also observe the proposed method produce accurate instance level segmentation on DAVIS-17 datasets.

We show the failure cases of our method in~\figref{fig:failure}. Possible reasons for our method to fail include tiny objects and similar appearance of different instances.

\section{Conclusion}
We present an efficient video object segmentation algorithm 
base on a novel soft matching layer. 
The method generalizes well and does not require online fine-tuning while maintaining good accuracy. 
Our method achieves state-of-the-art on the Youtube-Objects and JumpCut datasets and is competitive on DAVIS-16 and DAVIS-17, while its computational time  is at least one order of magnitude faster than current state-of-the-art. 

\noindent\textbf{Acknowledgments:} This material is based upon work supported in part by the National Science Foundation under Grant No.~1718221,~1755785, Samsung, and 3M. We thank NVIDIA for providing the GPUs used for this research.


\clearpage

\bibliographystyle{splncs04}
\bibliography{egbib,alex}

\begin{thebibliography}{10}
\providecommand{\url}[1]{\texttt{#1}}
\providecommand{\urlprefix}{URL }
\providecommand{\doi}[1]{https://doi.org/#1}

\bibitem{AvinashCVPR14}
Avinash~Ramakanth, S., Venkatesh~Babu, R.: Seam{S}eg: {V}ideo object
  segmentation using patch seams. In: Proc. CVPR (2014)

\bibitem{BaiSIGGRAPH2009}
Bai, X., Wang, J., Simons, D., Sapiro, G.: {Video snapcut: robust video object
  cutout using localized classifiers}. SIGGRAPH  (2009)

\bibitem{bertinetto2016fully}
Bertinetto, L., Valmadre, J., Henriques, J.F., Vedaldi, A., Torr, P.H.:
  Fully-convolutional siamese networks for object tracking. In: Proc. CVPR
  (2017)

\bibitem{caelles2017semantically}
Caelles, S., Chen, Y., Pont-Tuset, J., Van~Gool, L.: Semantically-guided video
  object segmentation. arXiv preprint arXiv:1704.01926  (2017)

\bibitem{caelles2016one}
Caelles, S., Maninis, K.K., Pont-Tuset, J., Leal-Taix{\'e}, L., Cremers, D.,
  Van~Gool, L.: One-shot video object segmentation. In: Proc. CVPR (2017)

\bibitem{ChenArxiv16}
Chen, L.C., Papandreou, G., Kokkinos, I., Murphy, K., Yuille, A.L.: Deeplab:
  Semantic image segmentation with deep convolutional nets, atrous convolution,
  and fully connected crfs. PAMI  (2018)

\bibitem{chen2018blazingly}
Chen, Y., Pont-Tuset, J., Montes, A., Van~Gool, L.: Blazingly fast video object
  segmentation with pixel-wise metric learning. In: Proc. CVPR (2018)

\bibitem{ChenICCV17}
Cheng, J., Tsai, Y.H., Wang, S., Yang, M.H.: {S}eg{F}low: Joint learning for
  video object segmentation and optical flow. In: Proc. ICCV (2017)

\bibitem{cheng2018fast}
Cheng, J., Tsai, Y.H., Hung, W.C., Wang, S., Yang, M.H.: Fast and accurate
  online video object segmentation via tracking parts. In: Proc. CVPR (2018)

\bibitem{dosovitskiy2015flownet}
Dosovitskiy, A., Fischer, P., Ilg, E., Hausser, P., Hazirbas, C., Golkov, V.,
  van~der Smagt, P., Cremers, D., Brox, T.: Flownet: Learning optical flow with
  convolutional networks. In: Proc. ICCV (2015)

\bibitem{EveringhamIJCV15}
Everingham, M., Eslami, S.A., Van~Gool, L., Williams, C.K., Winn, J.,
  Zisserman, A.: The pascal visual object classes challenge: A retrospective.
  IJCV  (2015)

\bibitem{FaktorBMVC14}
Faktor, A., Irani, M.: Video segmentation by non-local consensus voting. In:
  BMVC (2014)

\bibitem{FanTOG15}
Fan, Q., Zhong, F., Lischinski, D., Cohen-Or, D., Chen, B.: {J}ump{C}ut:
  {N}on-successive mask transfer and interpolation for video cutout. SIGGRAPH
  (2015)

\bibitem{GodecICCV2011}
Godec, M., Roth, P.M., Bischof, H.: {Hough-based tracking of non-rigid
  objects}. In: Proc. ICCV (2011)

\bibitem{GrundmannCVPR2010}
Grundmann, M., Kwatra, V., Han, M., Essa, I.: {Efficient hierarchical
  graph-based video segmentation}. In: Proc. CVPR (2010)

\bibitem{HariharanICCV11}
Hariharan, B., Arbel{\'a}ez, P., Bourdev, L., Maji, S., Malik, J.: Semantic
  contours from inverse detectors. In: Proc. ICCV (2011)

\bibitem{HeCVPR16}
He, K., Zhang, X., Ren, S., Sun, J.: Deep residual learning for image
  recognition. In: Proc. CVPR (2016)

\bibitem{Hu15}
Hu, Y.T., Lin, Y.Y., Chen, H.Y., Hsu, K.J., Chen, B.Y.: Matching images with
  multiple descriptors: An unsupervised approach for locally adaptive
  descriptor selection. TIP  (2015)

\bibitem{hu2017maskrnn}
Hu, Y.T., Huang, J.B., Schwing, A.: {MaskRNN}: Instance level video object
  segmentation. In: NIPS (2017)

\bibitem{hu2018unsupseg}
Hu, Y.T., Huang, J.B., Schwing, A.: Unsupervised video object segmentation
  using motion saliency-guided spatio-temporal propagation. In: Proc. ECCV
  (2018)

\bibitem{JainECCV2014}
Jain, S.D., Grauman, K.: {Supervoxel-consistent foreground propagation in
  video}. In: Proc. ECCV (2014)

\bibitem{jain2017fusionseg}
Jain, S.D., Xiong, B., Grauman, K.: {F}usion{S}eg: Learning to combine motion
  and appearance for fully automatic segmention of generic objects in videos.
  Proc. CVPR  (2017)

\bibitem{JampaniCVPR17}
Jampani, V., Gadde, R., Gehler, P.V.: Video propagation networks. In: Proc.
  CVPR (2017)

\bibitem{JangCVPR17}
Jang, W.D., Kim, C.S.: Online video object segmentation via convolutional
  trident network. In: Proc. CVPR (2017)

\bibitem{PerazziCVPR17}
Khoreva, A., Perazzi, F., Benenson, R., Schiele, B., A.Sorkine-Hornung:
  Learning video object segmentation from static images. In: Proc. CVPR (2017)

\bibitem{khoreva2017lucid}
Khoreva, A., Benenson, R., Ilg, E., Brox, T., Schiele, B.: Lucid data dreaming
  for object tracking. arXiv preprint arXiv:1703.09554  (2017)

\bibitem{kingma2014adam}
Kingma, D., Ba, J.: Adam: A method for stochastic optimization. In:
  International Conference on Learning Representations (2014)

\bibitem{VOT_TPAMI}
Kristan, M., Matas, J., Leonardis, A., Vojir, T., Pflugfelder, R., Fernandez,
  G., Nebehay, G., Porikli, F., \v{C}ehovin, L.: A novel performance evaluation
  methodology for single-target trackers. PAMI  (2016)

\bibitem{LeeICCV11}
Lee, Y.J., Kim, J., Grauman, K.: Key-segments for video object segmentation.
  In: Proc. ICCV (2011)

\bibitem{LezamaCVPR11}
Lezama, J., Alahari, K., Sivic, J., Laptev, I.: Track to the future:
  {S}patio-temporal video segmentation with long-range motion cues. In: Proc.
  CVPR (2011)

\bibitem{LiICCV13}
Li, F., Kim, T., Humayun, A., Tsai, D., Rehg, J.M.: Video segmentation by
  tracking many figure-ground segments. In: Proc. ICCV (2013)

\bibitem{LiCVPRW2017}
Li, X., Qi, Y., Wang, Z., Chen, K., Liu, Z., Shi, J., Luo, P., Loy, C.C., Tang,
  X.: Video object segmentation with re-identification. The 2017 DAVIS
  Challenge on Video Object Segmentation - CVPR Workshops  (2017)

\bibitem{Lowe04}
Lowe, D.: Distinctive image features from scale-invariant keypoints. IJCV
  (2004)

\bibitem{MaerkiCVPR16}
Maerki, N., Perazzi, F., Wang, O., Sorkine-Hornung, A.: Bilateral space video
  segmentation. In: Proc. CVPR (2016)

\bibitem{Mikolajczyk05b}
Mikolajczyk, K., Schmid, C.: A performance evaluation of local descriptors.
  IEEE Transactions on Pattern Analysis and Machine Intelligence  (2005)

\bibitem{NagarajaICCV15}
Nagaraja, N., Schmidt, F., Brox, T.: Video segmentation with just a few
  strokes. In: Proc. ICCV (2015)

\bibitem{OchsPAMI2014}
Ochs, P., Malik, J., Brox, T.: {Segmentation of moving objects by long term
  video analysis}. PAMI  (2014)

\bibitem{oh2018fast}
Oh, S.W., Lee, J.Y., Sunkavalli, K., Kim, S.J.: Fast video object segmentation
  by reference-guided mask propagation. In: Proc. CVPR (2018)

\bibitem{PapazoglouICCV13}
Papazoglou, A., Ferrari, V.: Fast object segmentation in unconstrained video.
  In: Proc. ICCV (2013)

\bibitem{PerazziCVPR16}
Perazzi, F., Pont-Tuset, J., McWilliams, B., Gool, L.V., Gross, M.,
  Sorkine-Hornung, A.: A benchmark dataset and evaluation methodology for video
  object segmentation. In: Proc. CVPR (2016)

\bibitem{PerazziICCV15}
Perazzi, F., Wang, O., Gross, M., Sorkine-Hornung, A.: Fully connected object
  proposals for video segmentation. In: Proc. ICCV (2015)

\bibitem{pont2017DAVIS}
Pont-Tuset, J., Perazzi, F., Caelles, S., Arbel{\'a}ez, P., Sorkine-Hornung,
  A., Van~Gool, L.: The 2017 davis challenge on video object segmentation.
  arXiv preprint arXiv:1704.00675  (2017)

\bibitem{PrestCVPR12}
Prest, A., Leistner, C., Civera, J., Schmid, C., Ferrari, V.: Learning object
  class detectors from weakly annotated video. In: Proc. CVPR (2012)

\bibitem{PriceICCV09}
Price, B.L., Morse, B.S., Cohen, S.: {LIVE}cut: Learning-based interactive
  video segmentation by evaluation of multiple propagated cues. In: Proc. ICCV
  (2009)

\bibitem{Revaud2016}
Revaud, J., Weinzaepfel, P., Harchaoui, Z., Schmid, C.: Deepmatching:
  Hierarchical deformable dense matching. IJCV  (2016)

\bibitem{rocco2017convolutional}
Rocco, I., Arandjelovic, R., Sivic, J.: Convolutional neural network
  architecture for geometric matching. In: Proc. CVPR (2017)

\bibitem{tokmakov2017learning1}
Tokmakov, P., Alahari, K., Schmid, C.: Learning motion patterns in videos. In:
  Proc. CVPR (2017)

\bibitem{tokmakov2017learning2}
Tokmakov, P., Alahari, K., Schmid, C.: Learning video object segmentation with
  visual memory. In: Proc. ICCV (2017)

\bibitem{TsaiBMVC2010}
Tsai, D., Flagg, M., Rehg, J.: {Motion coherent tracking with multi-label mrf
  optimization}. In: Proc. BMVC (2010)

\bibitem{TsaiCVPR2016}
Tsai, Y.H., Yang, M.H., Black, M.J.: {Video Segmentation via Object Flow}. In:
  Proc. CVPR (2016)

\bibitem{VijayanarasimhanECCV2012}
Vijayanarasimhan, S., Grauman, K.: {Active frame selection for label
  propagation in videos}. In: Proc. ECCV (2012)

\bibitem{DAVIS2017_5th}
Voigtlaender, P., Leibe, B.: Online adaptation of convolutional neural networks
  for the 2017 davis challenge on video object segmentation. The 2017 DAVIS
  Challenge on Video Object Segmentation - CVPR Workshops  (2017)

\bibitem{voigtlaender2017online}
Voigtlaender, P., Leibe, B.: Online adaptation of convolutional neural networks
  for video object segmentation. BMVC  (2017)

\bibitem{WangTIP2017}
Wang, W., Shen, J., Porikli, F.: Selective video object cutout. TIP  (2017)

\bibitem{XiaoCVPR16}
Xiao, F., Lee, Y.J.: Track and segment: An iterative unsupervised approach for
  video object proposals. In: Proc. CVPR (2016)

\bibitem{yang2018efficient}
Yang, L., Wang, Y., Xiong, X., Yang, J., Katsaggelos, A.K.: Efficient video
  object segmentation via network modulation. In: Proc. CVPR (2018)

\bibitem{yang2017deepcd}
Yang, T.Y., Hsu, J.H., Lin, Y.Y., Chuang, Y.Y.: Deepcd: Learning deep
  complementary descriptors for patch representations. In: Proc. ICCV (2017)

\bibitem{yilmaz2006object}
Yilmaz, A., Javed, O., Shah, M.: Object tracking: A survey. Acm computing
  surveys (CSUR)  (2006)

\bibitem{yoon2017pixel}
Yoon, J.S., Rameau, F., Kim, J., Lee, S., Shin, S., Kweon, I.S.: Pixel-level
  matching for video object segmentation using convolutional neural networks.
  In: Proc. ICCV (2017)

\bibitem{ZhongTOG12}
Zhong, F., Qin, X., Peng, Q., Meng, X.: Discontinuity-aware video object
  cutout. SIGGRAPH  (2012)

\end{thebibliography}
\end{document}